%% file: main.tex
\documentclass[conference]{IEEEtran}
\usepackage{times}

\usepackage[numbers]{natbib}
\usepackage{multicol}

\usepackage{xspace}
\usepackage[colorlinks=true,linkcolor=blue,citecolor=blue,bookmarks=true]{hyperref}

\usepackage{wrapfig}

\usepackage{graphicx}
\usepackage{subcaption}
\usepackage{booktabs}

\usepackage[skins,theorems]{tcolorbox}

\include{defs}

\include{math_commands}

\usepackage{titlesec}
\titlespacing\section{0pt}{0pt plus 2pt minus 2pt}{0pt plus 2pt minus 2pt}
\titlespacing\subsection{0pt}{3pt plus 4pt minus 2pt}{0pt plus 2pt minus 2pt}
\titlespacing\subsubsection{0pt}{3pt plus 4pt minus 2pt}{0pt plus 2pt minus 2pt}

\begin{document}

\title{Pre-Training for Robots: Offline RL Enables Learning New Tasks in a Handful of Trials}

\author{Aviral Kumar$^{*, 1}$, Anikait Singh$^{*, 1}$, Frederik Ebert$^{*, 1}$, Mitsuhiko Nakamoto$^1$, Yanlai Yang$^3$, \\ Chelsea Finn$^2$, Sergey Levine$^1$ \vspace{2pt}\\
\small{$^1$UC Berkeley, $^2$Stanford University, $^3$New York University}~~~~~ ($^*$Equal contribution)
}

\makeatletter
\let\@oldmaketitle\@maketitle%
\renewcommand{\@maketitle}{\@oldmaketitle%
    \centering
  \includegraphics[width=0.9\linewidth]{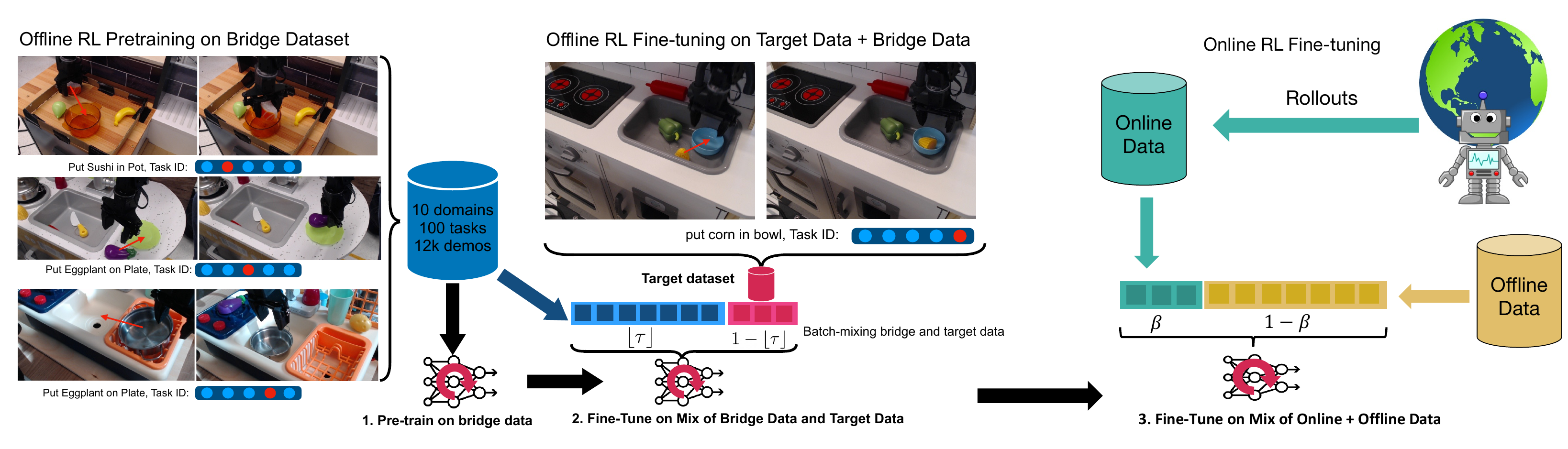}
  \vspace{-0.2cm}
  \captionof{figure}{ \label{fig:system_overview} \footnotesize \textbf{Overview of \methodname:} We first perform general offline pre-training on diverse multi-task robot data and subsequently fine-tune on one or several target tasks while mixing batches between the prior data and the target dataset using a batch mixing ratio of $\tau$. Additionally, a separate online fine-tuning phase can be done, where offline pre-training is done on a static dataset and an online replay buffer is collected using rollouts in the environment. The offline and online buffers are mixed per batch with a ratio of $\beta$.}
  \vspace{-0.35cm}
 }
\makeatother

\maketitle
\pagestyle{empty}

\begin{abstract}
Progress in deep learning highlights the tremendous potential of utilizing diverse robotic datasets for attaining effective generalization and makes it enticing to consider leveraging broad datasets for attaining robust generalization in robotic learning as well. However, in practice, we often want to learn a new skill in a new environment that is unlikely to be contained in the prior data. Therefore we ask: how can we leverage existing diverse offline datasets in combination with small amounts of task-specific data to solve new tasks, while still enjoying the generalization benefits of training on large amounts of data? In this paper, we demonstrate that end-to-end offline RL can be an effective approach for doing this, without the need for any representation learning or vision-based pre-training. We present pre-training for robots (PTR), a framework based on offline RL that attempts to effectively learn new tasks by combining pre-training on existing robotic datasets with rapid fine-tuning on a new task, with as few as 10 demonstrations. PTR utilizes an existing offline RL method, conservative Q-learning (CQL), but extends it to include several crucial design decisions that enable PTR to actually work and outperform a variety of prior methods. To our knowledge, PTR is the first RL method that succeeds at learning new tasks in a new domain on a real WidowX robot with as few as 10 task demonstrations, by effectively leveraging an existing dataset of diverse multi-task robot data collected in a variety of toy kitchens. We also demonstrate that PTR can enable effective autonomous fine-tuning and improvement in a handful of trials, without needing any demonstrations. An accompanying overview video can be found in the supplementary material and at this URL: \url{https://sites.google.com/view/ptr-final/}
\end{abstract}

\IEEEpeerreviewmaketitle

\input{introduction}


\input{related}
	

\input{prelims}

\input{method}

\input{experiments}

\vspace{0.1cm}
\section{Discussion and Conclusion}
\label{sec:conclusion}
\vspace{0.1cm}
We presented a system that uses diverse prior data for general-purpose offline RL pre-training, followed by fine-tuning to downstream tasks. The prior data, sourced from a publicly available dataset, consists of over a hundred tasks across ten scenes and our policies can be fine-tuned with as few as 10 demonstrations. We show that this approach outperforms prior pre-training and fine-tuning methods based on imitation learning. One of the most exciting directions for future work is to further scale up this pre-training to provide a single policy initialization, that can be utilized as a starting point, similar to GPT3~\citep{brown2020language}. 
An exciting future direction is to scale PTR up to more complex settings, including to novel robots. {Since joint training with offline RL was worse than pre-training and then fine-tuning with PTR, another exciting direction for future work is to understand the pros and cons of joint training and fine-tuning in the context of robot learning.}

\vspace{0.1cm}







\bibliography{main}
\bibliographystyle{plainnat}

\newpage
\clearpage
\appendix

\subsection{Diagnostic study in simulation}
\label{app:sim_diagnostic}
 We perform a diagnostic study in simulation to verify some of the insights observed in our real-world experiments. We created a bin sort task, where a WidowX250 robot is placed in front two bins and is provided with two objects (more details in Appendix~\ref{app:exp_setup}). The task is to sort each object in the correct bin associated with that object. The pre-training data provided to this robot is pick-place data, which only demonstrates how to pick \emph{one} of the objects and place it in one of the bins, but does not demonstrate the compound task of placing both objects. In order to succeed at this such a compound task, a robot must learn an abstract representation of the skill of sorting an object during the pre-training phase and then figure out that it needs to apply this skill multiple times in a trajectory to succeed at the task from just \emph{five} demonstrations of the desired sorting behavior.

The performance numbers (along with 95\%-confidence intervals) are shown in Table~\ref{tab:sim_complete}. Observe that \methodname improves upon prior methods in a statistically significant manner, outperforming the BC and COG baselines by a significant margin. This validates the efficacy of \methodname in simulation and corroborates our real-world results.

\begin{table}[h]
\centering
\begin{tabular}{l|r}
\toprule
\textbf{Method} & \textbf{Success rate}  \\ \midrule
BC (joint training) & 7.00 $\pm$ 0.00 \% \\
COG (joint training) & 8.00 $\pm$ 1.00 \% \\
BC (finetune) & 4.88 $\pm$ 4.07 \% \\ \midrule
\textbf{\methodname (Ours)} & \textbf{17.41 $\pm$ 1.77 \%} \\
\bottomrule
\end{tabular}
\caption{\label{tab:sim_complete} \footnotesize{\textbf{Performance of \methodname in comparison with other methods} on the simulated bin sorting task, trained for many more gradient steps for all methods until each one of them converges. Observe that \methodname substantially outperforms other prior methods, including joint training on the same data with BC or CQL. Training on target data only is unable to recover a non-zero performance, so we do not report it in this table. Since the 95\%-confidence intervals do not overlap between \methodname and other methods, it indicates that \methodname improves upon baselines in a statistically significant manner.}}
\end{table}

\subsection{Details of Our Experimental Setup}
\label{app:exp_setup}

\vspace{0.1cm}
\subsubsection{Real-World Experimental Setup}

\begin{figure}[h]
\centering
  \includegraphics[width=\linewidth]{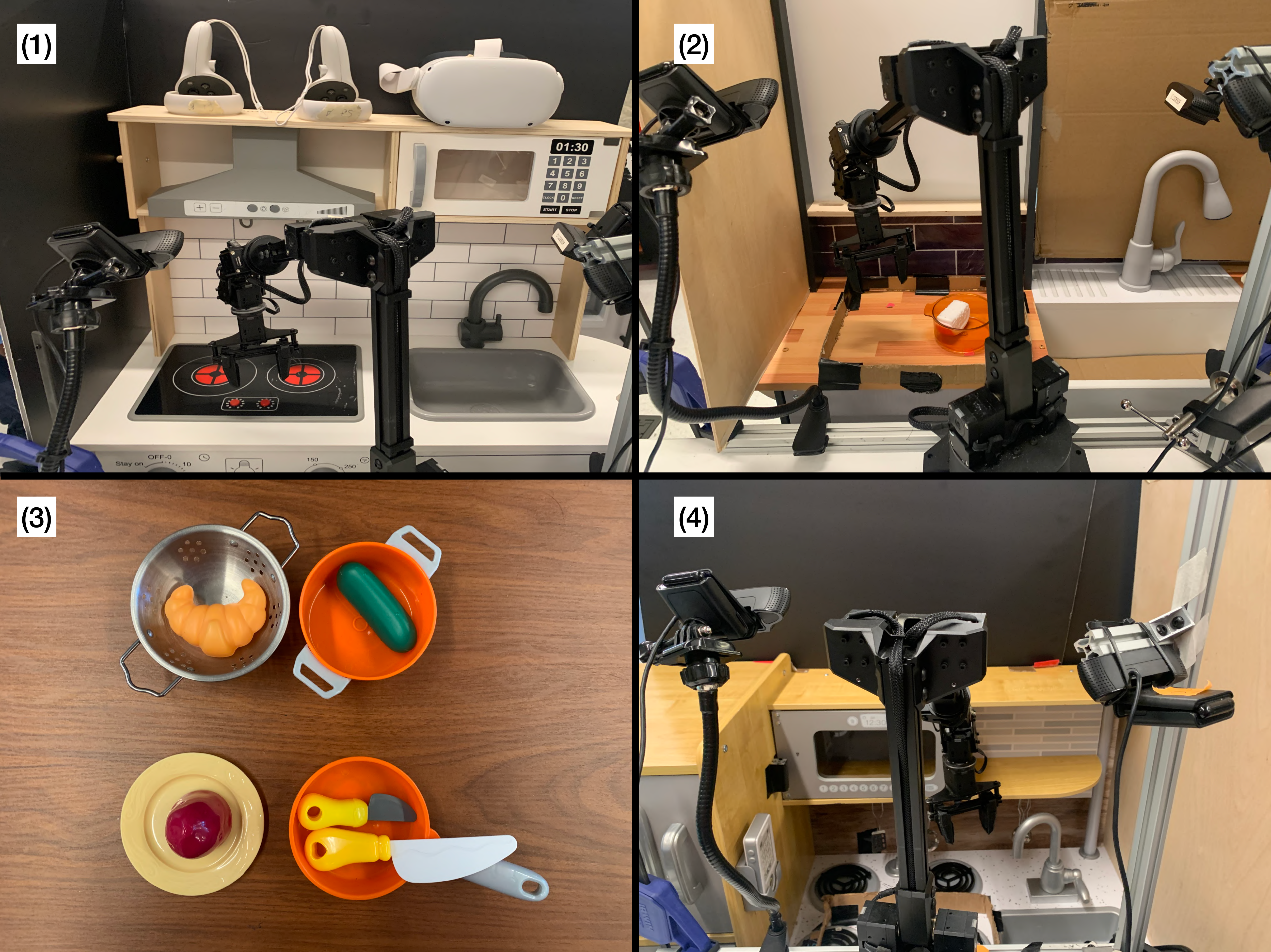}
  \caption{\footnotesize{\textbf{Setup Overview}: Following \citet{ebert2021bridge}, we use a toykitchen setup described in that prior work for our experiments. This utilizes a 6-DoF WidowX 250 robot. \textbf{(1):}  Held-out toykitchen used for experiments in Scenario 3 (denoted ``toykitchen 6''), \textbf{(2):}  Re-targeting toykitchen used for experiments in Scenario 2 (denoted ``toykitchen 2''), \textbf{(3):} target objects used in the experiments of scenario 3.}, \textbf{(4):} the held-out kitchen setup used for door opening (``toykitchen 1'').}
  \label{fig:setup_overview}
  \vspace{-0.3cm}
\end{figure}

A picture of our real-world experimental setup is shown in Figure \ref{fig:setup_overview}. The scenarios considered in our experiments (Section~\ref{sec:result}) are designed to evaluate the performance of our method under a variety of situations and therefore we set up these tasks in different toykitchen domains (see Figure \ref{fig:setup_overview}) on three different WidowX 250 robot arms. We use data from the bridge dataset~\citep{ebert2021bridge} consisting of data collected with many robots in many domains for training but exclude the task/domain that we use for evaluation from the training dataset.  

\subsubsection{Diagnostic Experimental Setup in Simulation}
\label{sec:sim_appendix}

\begin{figure}{}
\centering
    \includegraphics[width=0.8\linewidth]{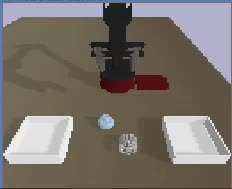}
  \caption{\footnotesize{\textbf{Bin-Sorting task used for our simulated evaluations.} The task requires sorting the cylinder into the left bin and the teapot into the right bin.}}
  \label{app:sim_setup}
\end{figure}

We evaluate our approach in a simulated bin-sorting task on the simulated WidowX 250 platform, aimed to mimic the setup we use for our real-world evaluations. This setup is designed in the PyBullet simulation framework provided by \citet{singh2020cog}. A picture is shown in Figure \ref{app:sim_setup}. In this task, two different bins and two different objects are placed in front of the WidowX robot. The goal of the robot is to correctly sort each of the two objects to their designated bin (e.g the cylinder is supposed to be placed in the left bin and the teapot should be placed in the right bin. We refer to this task as a \emph{compound} task since it requires successfully combining behaviors of two different pick-and-place skills one after the other in a single trajectory while also adequately identifying the correct bin associated with each object. Success is counted only when the robot can accurately sort \emph{both} of the objects into their corresponding bins.

\begin{figure*}
\centering
  \includegraphics[width=\textwidth]{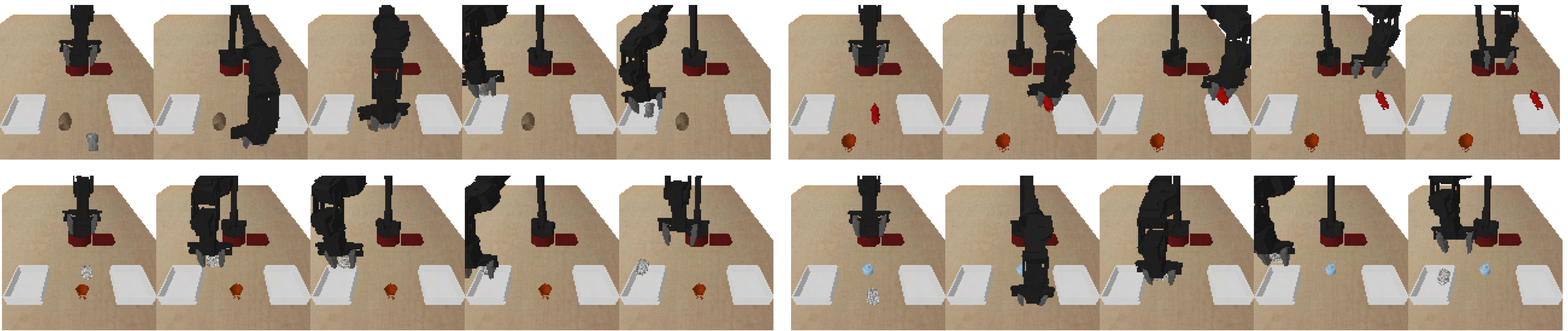}
  \caption{\label{fig:app_pretrain_rollout_sim} \footnotesize {Some trajectories from the pre-training data used in the simulated bin-sort task.}}
\end{figure*}

\begin{figure*}
\centering
  \includegraphics[width=\textwidth]{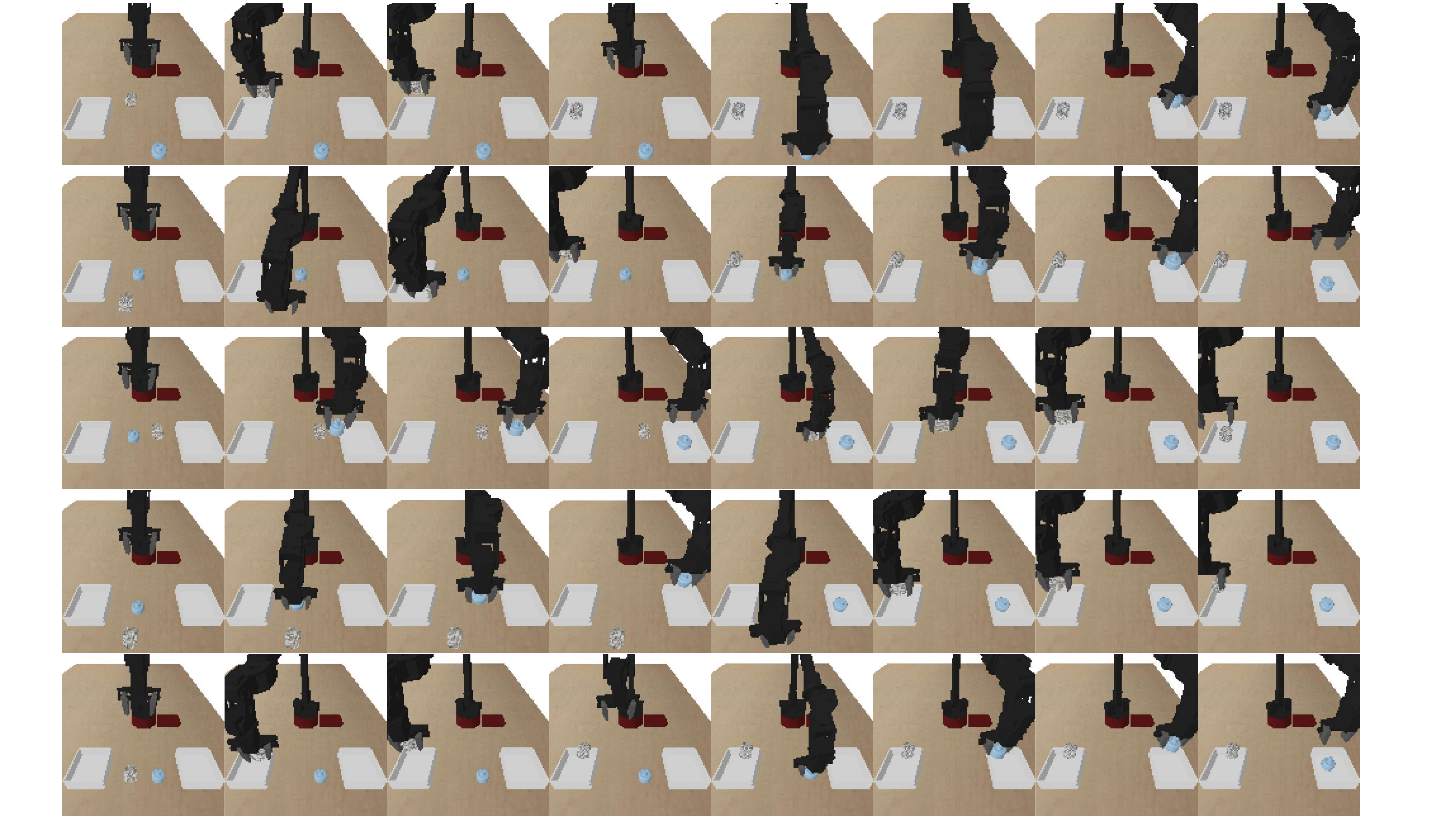}
  \caption{\label{fig:app_targ_rollout_sim} \footnotesize {The five demonstration trajectories used for Phase 2 of \methodname.}}
\end{figure*}

\textbf{Offline pre-training dataset.} The dataset provided for offline pre-training only consists of demonstrations that show how the robot should pick one of the two objects and place it into one of the two bins. Each episode in the pre-training dataset is about 30-40 timesteps long. A picture showing some trajectories from the pre-training dataset is shown in Figure \ref{fig:app_pretrain_rollout_sim}. While the downstream task only requires solving this sorting task with two specific objects (shown in Figure \ref{fig:app_targ_rollout_sim}), the pre-training data consists of 10 unique objects (some shown in Figure \ref{fig:app_pretrain_rollout_sim}). The two target objects that appear together in the downstream target scene are never seen together in the pre-training data. Since the pre-training data only demonstrates how the robot must pick up one of the objects and place it in one of the two bins (not necessarily in the target bin that the target task requires), it neither consists of any behavior that places objects into bins sequentially nor does it consist of any behavior where one of the objects is placed one of the bins while the other one is not. This is what makes this  task particularly challenging.

\textbf{Target demonstration data.} The target task data provided to the algorithm consists of only \textbf{\emph{five}} demonstrations that show how the robot must complete both the stages of placing both objects (see Figure \ref{fig:app_targ_rollout_sim}). Each episode in the target demonstration data is 80 timesteps long, which is substantially longer than any trajectory in the pre-training data, though one would hope that good representations learned from the pick and place tasks are still useful for this target task. While all methods are able to generally solve the first segment of placing the first object into the correct bin, the primary challenge in this task is to effectively sort the second object, and we find that \methodname attains a substantially better success rate than other baselines in this exact step.  

\subsection{Description of the Real-World Evaluation Scenarios}
\label{app:tasks}
In this section, we describe the real-world evaluation scenarios considered in Section~\ref{sec:result}. We additionally include a much more challenging version of Scenario 3, for which we present results in Appendix~\ref{app:exp_results}. These harder test cases evaluate the fine-tuning performance on four different tasks, starting from the same initialization trained on bridge data except the toykitchen 6 domain in which these four tasks were set up. In the following sections, the nomenclature for the toy kitchens is drawn from \citet{ebert2021bridge} and as described in the caption of Figure~\ref{fig:setup_overview}.

\subsubsection{Scenario 1: Re-targeting skills for existing to solve new tasks}

\begin{figure}
\centering
  \includegraphics[width=0.83\linewidth]{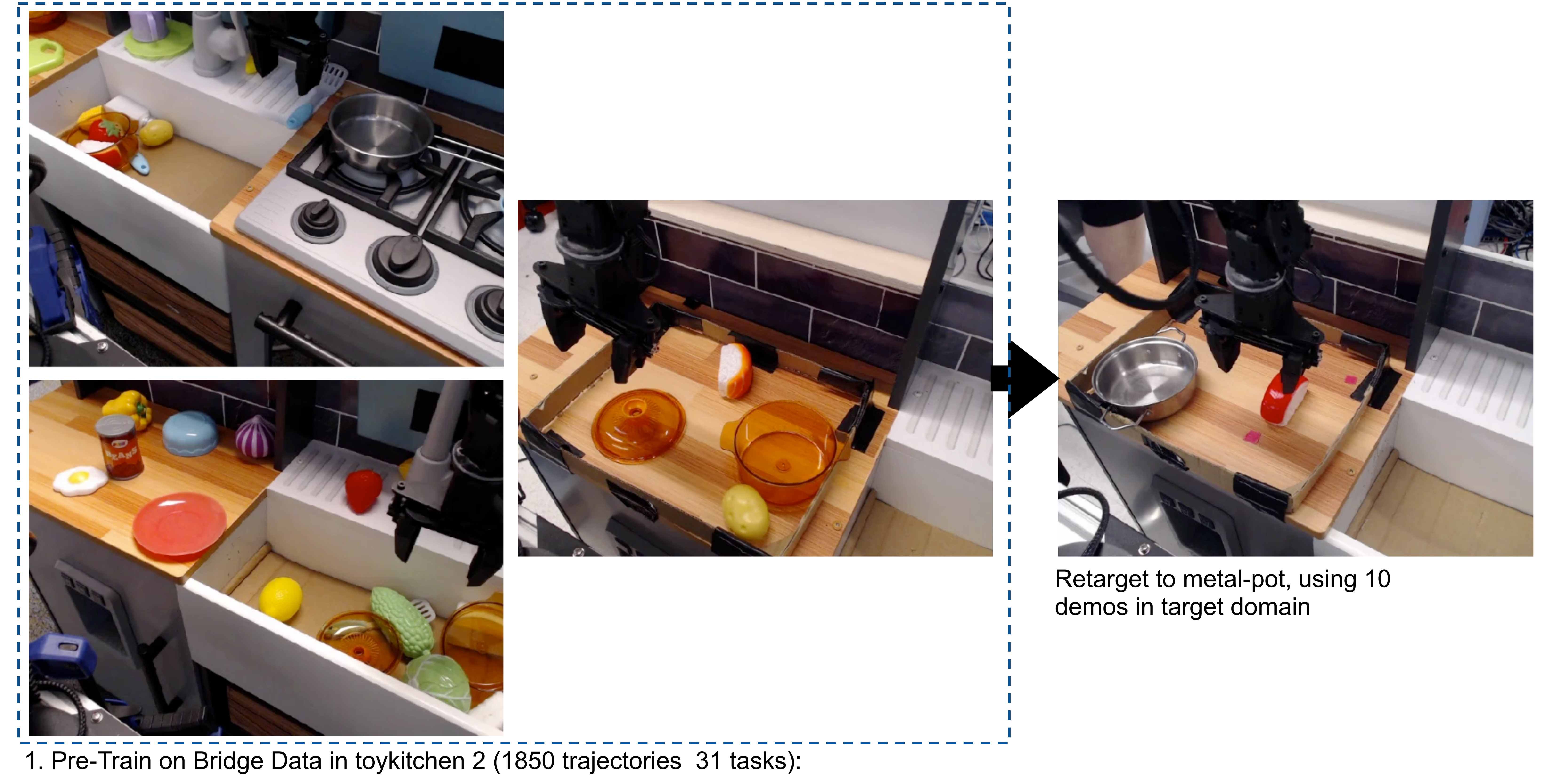}
  \caption{\footnotesize \textbf{Illustration of pre-training data and finetuning data used for Scenario 1}: re-targeting the put sushi in metal-pot behavior to put the object in the metal pot instead of the orange transparent pot.}
  \label{fig:retargeting_setup}
\end{figure}

\textbf{Pre-training data.} The pre-training data comprises all of the pick and place data from the bridge dataset~\citep{ebert2021bridge} from toykitchen 2. This includes data corresponding to the task of putting the sushi in the transparent orange pot (Figure \ref{fig:retargeting_setup}).  

\textbf{Target task and data.} Since our goal in this scenario is to re-target the skill for putting the sushi in the transparent orange pot to the task of putting the sushi in the metallic pot, we utilize a dataset of 20 demonstrations that place the sushi in a metallic pot as our target task data that we fine-tune with (shown in Figure~\ref{fig:retargeting_setup}). 

\textbf{Quantitative evaluation protocol.} For our quantitative evaluations in Table \ref{tab:retarget}, we run 10 controlled evaluation rollouts that place the sushi and the metallic pot in different locations of the workspace. In all runs, the arm starts at up to 10 cm distance above the target object. The initial object and arm poses and positions are matched as closely as possible for different methods.

\subsubsection{Scenario 2: Generalizing to Previously Unseen Domains}

\begin{figure}
\centering
  \includegraphics[width=0.83\linewidth]{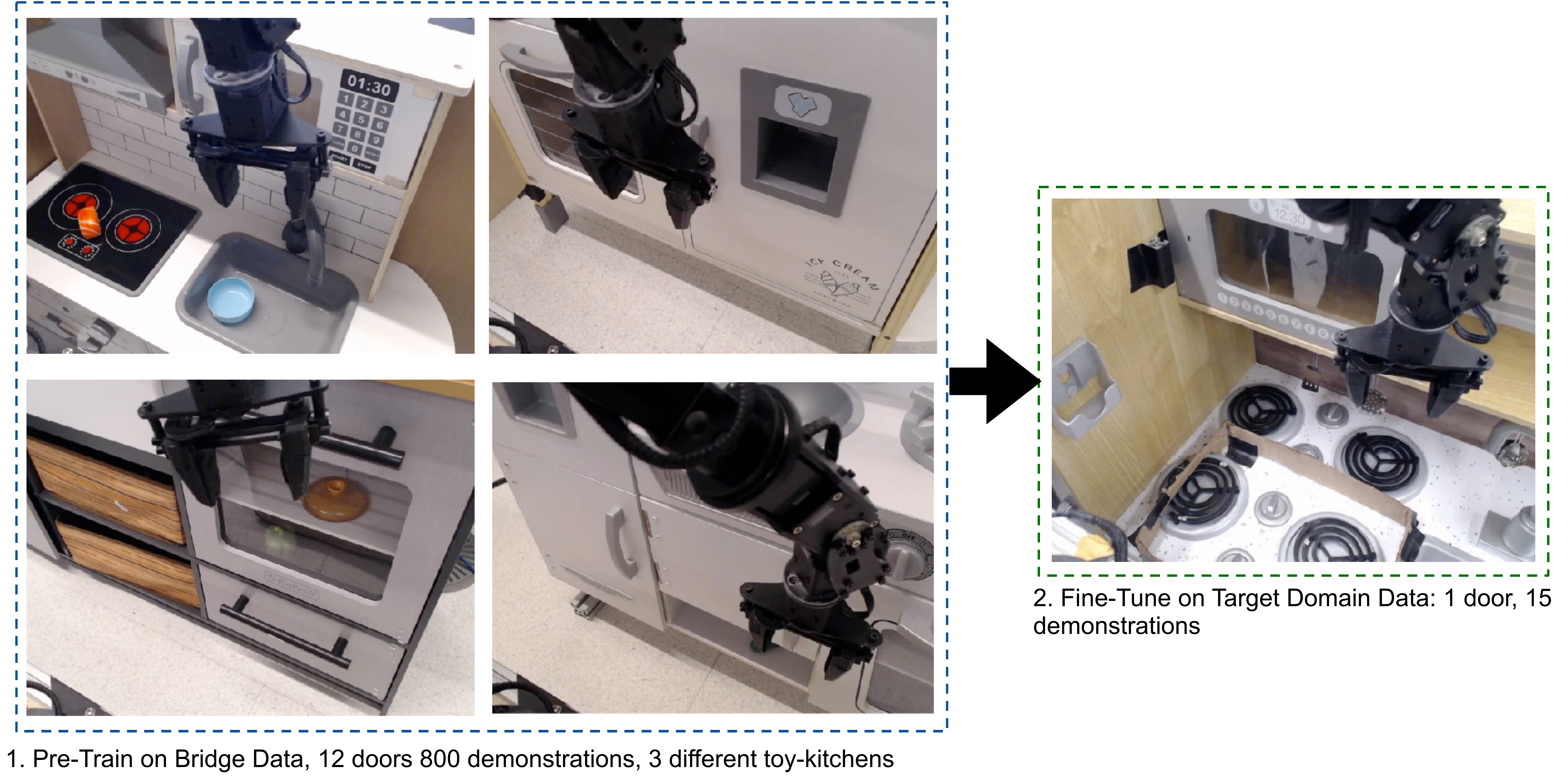}
  \caption{\footnotesize \textbf{Illustration of pre-training data and fine-tuning data used for Scenario 2 (door opening)}: transferring a behavior to a held-out domain.}
  \vspace{-0.5cm}
  \label{fig:door_open_setup}
\end{figure}

\textbf{Pre-training data.} The pre-training data in Scenario 2 consists of 800 door-opening demonstrations on 12 different doors across 3 different toykitchen domains.

\textbf{Target task and data.} The target task requires opening the door of an unseen microwave in toykitchen 1 using a target dataset of only 15 demonstrations.

\textbf{Quantitative evaluation protocol.} We run 20 rollouts with each method, counting successes when the robot opened the door by at least 45 degrees. To perform this successfully, there is a degree of complexity as the robot has to initially open the door till it's open to about 30 degrees. Then due to physical constraints, the robot needs to wrap around the door and push it open from the inside. To begin an evaluation rollout, we reset the robot to randomly sampled poses obtained from held-out demonstrations on the target door.  This is a compound task requiring the robot to first grab the door by the handle, next move around the door, and finally push the door open. As before, we match the initial pose of the robot as closely as possible for all the methods.

\subsubsection{Scenario 3: Learning to Solve New Tasks in New Domains}

\begin{figure}
\centering
  \includegraphics[width=0.83\linewidth]{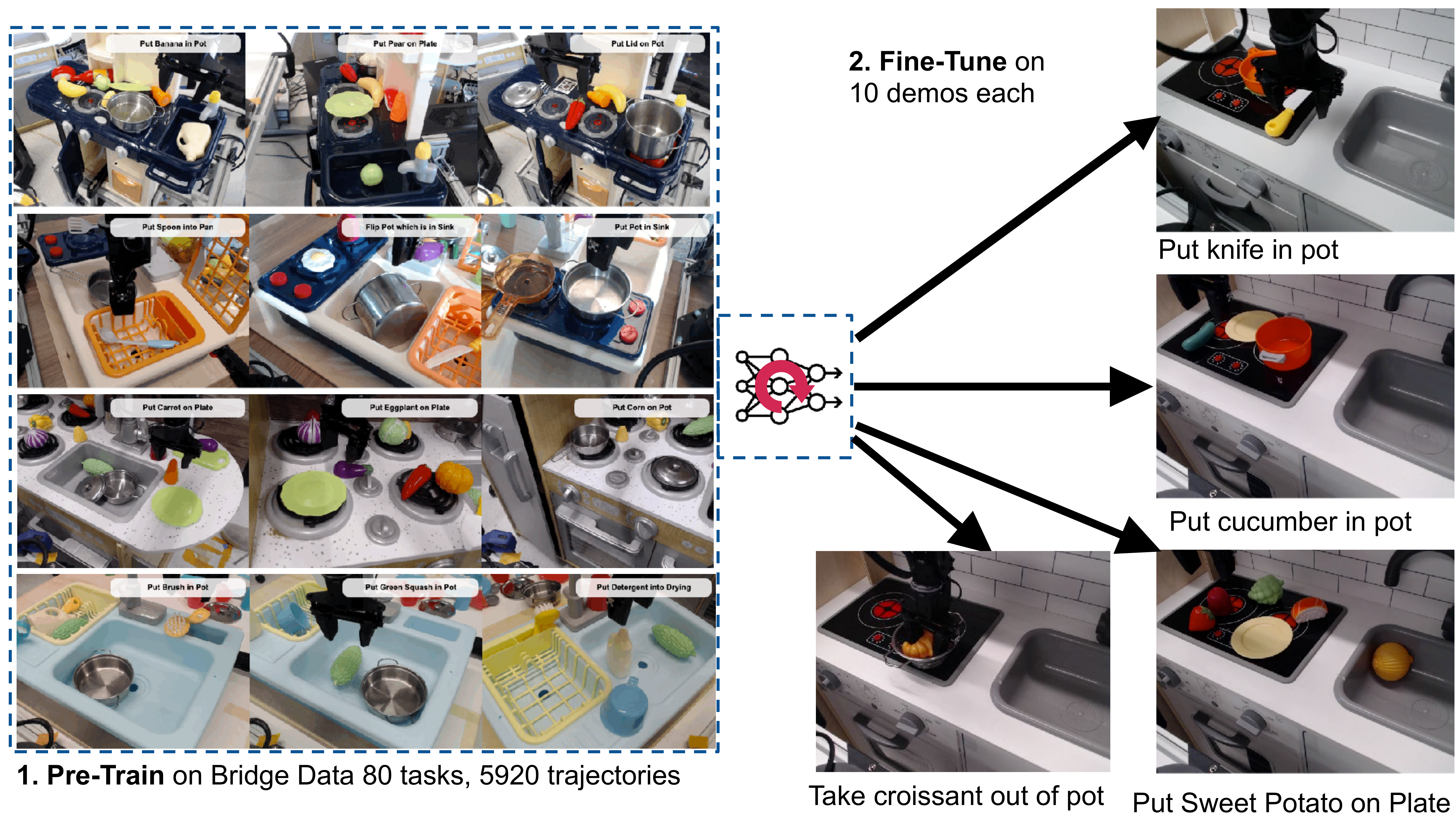}
  \caption{\footnotesize {\textbf{Illustration of pre-training data and fine-tuning data used for the new tasks we have added in Scenario 3}. The goal is to learn to solve new tasks in new domains starting from the same pre-trained initialization and when fine-tuning is only performed using 10-20 demonstrations of the target task.}}
  \label{fig:scenario4_overview}
\end{figure}

\textbf{Pre-training data.} All pick-and-place data in the bridge dataset~\citep{ebert2021bridge} except any demonstration data collected in toykitchen~6, where our evaluations are performed.

\textbf{Target task and data.} The target task requires placing corn in a pot in the sink in the new target domain and the target dataset provides 10 demonstrations for this task. These target demonstrations are sampled from the bridge dataset itself.

\textbf{Quantitative evaluation protocol.} During the evaluation we were unable to exactly match the camera orientation used to collect the target demonstration trajectories, and therefore ran evaluations with a slightly modified camera view. This presents an additional challenge for any method as it must now generalize to a modified camera view of the target toykitchen domain, without having ever observed this domain or this camera view during training. We sampled initial poses for our method by choosing transitions from a held-out dataset of demonstrations of the target task and resetting the robot to those initial poses for each method. We attempted to match the positions of objects across methods as closely as possible.

\subsubsection{More Tasks in Scenario 3: Learning to Solve Multiple New Tasks in New Domains From the Same Initialization}
\label{app:scenario4}

In Appendix~\ref{app:exp_results}, we have now added results for more tasks in Scenario 3. The details of these tasks are as follows:

\textbf{Pre-training data.} All pick-and-place data from bridge dataset~\citep{ebert2021bridge} except data from toykitchen~6.

\textbf{Target task and data.} We consider four downstream tasks: take croissant from a metallic bowl, put sweet potato on a plate, place the knife in a pot, and put cucumber in a bowl. We collected 10 target demonstrations for the croissant, sweet potato, and put cucumber in bowl tasks, and 20 target demonstrations for the knife in pot task. A picture of these target tasks is shown in Figure \ref{fig:scenario4_overview}.

\textbf{Qualitative evaluation protocol.} For our evaluations, we utilize either 10 or 20 evaluation rollouts. As with all of our other quantitative results, we evaluate all the baseline approaches and \methodname starting from an identical set of initial poses for the robot. These initial poses are randomly sampled from the poses that appear in the first 10 timesteps of the held-out demonstration trajectories for this target task. For the configuration of objects, we test our policies in a variety of task-specific configurations that we discuss below:

\begin{itemize}
    \item \textbf{Take croissant from metallic bowl:} For this task, we alternate between two kinds of positions for the metallic bowl. In the ``easy'' positions, the metallic bowl is placed roughly vertically beneath the robot's initial starting pose, whereas in the ``hard'' positions, the robot must first move itself to the right location of the bowl and then execute the policy.
    \item \textbf{Put the cucumber in bowl:} We run 10 evaluation rollouts starting from 10 randomly sampled initial poses of the robot for our evaluations. Here we moved the bowl between the two stovetops in each trial. 
    \item \textbf{Put sweet potato on plate:} For this task, we performed 20 evaluation rollouts. We only sampled 10 initial poses for the robot, but for each position, we evaluated every policy on two orientations of the sweet potato (i.e., the sweet potato is placed on the table on its flat face or on its curved face). Each of these orientations presents some unique challenges, and evaluating both of them allows us to gauge how robust the learned policy is to changes in orientation. The demonstration data had a variety of orientations for the sweet potato object that differed for each collected trajectory. 
    \item \textbf{Place knife in pot:} We evaluate this task over 10 evaluation rollouts, where the first five rollouts use a smaller knife, while the other five rollouts use a larger knife (shown in Figure~\ref{fig:setup_overview}). Each knife was seen in the demonstration dataset with equal probability.
\end{itemize}

We will discuss the results obtained on these new tasks in Appendix~\ref{app:exp_results}.

\subsection{Additional Experimental Results}
\label{app:exp_results}

\begin{figure}[h]
\vspace{-0.3cm}
\centering
  \includegraphics[width=0.9\linewidth]{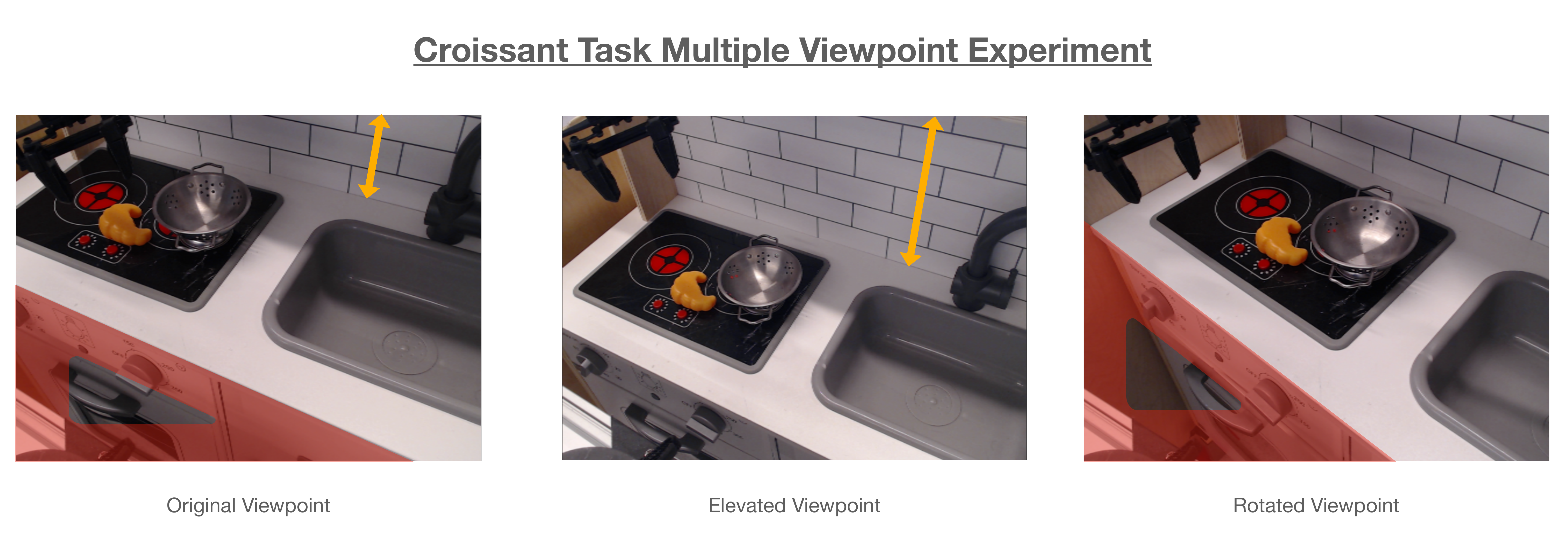}
  \vspace{-0.3cm}
  \caption{\footnotesize \textbf{Sample observations from different camera viewpoints, only used during fine-tuning}. \textbf{Left:} the original camera viewpoint found in Figure~\ref{fig:scenario4_overview}. \textbf{Middle:} an elevated camera viewpoint where the robot and camera have been raised 7 cm. \textbf{Right:} a rotated camera viewpoint where the kitchen has been slightly translated and rotated 15 degrees counterclockwise relative to the camera and robot.}
  \label{fig:multviewpoint}
  \vspace{-0.2cm}
\end{figure}

\textbf{Finetuning to novel camera viewpoints:} Even though Scenario 3 already presents a novel toy-kitchen domain and previously unseen objects during finetuning, we also evaluate \methodname on a more challenging scenario where we additionally alter the camera viewpoint during finetuning. We apply two kinds of alterations to the camera: \textbf{(a)} we elevate the mounting platform of the camera by 7 cm, which necessitates adapting the way the physical coordinates of the robot end-effector are interpreted by the policy, and \textbf{(b)} we rotate the camera by about 15 degrees to induce a more oblique image observation than what was ever seen during pre-training. Note that in both of these scenarios, the robot has never encountered such camera viewpoints during pre-training, which makes this scenario even more challenging. The original dataset in \citep{ebert2021bridge} had the camera elevated to the same position for each domain and always ensured the kitchen was parallel to the camera platform, with translations being the primary changes in the scene for each domain. In Table \ref{tab:viewpoint_comp}, we present our results comparing \methodname and BC (finetune). Observe that \methodname still clearly outperforms BC (finetune), and attains performance close to that of \methodname in Table~\ref{tab:scenario4}, indicating that such shifts in the camera do not drastically hurt \methodname.

\begin{table}[h]
\centering
\begin{tabular}{l|r|r}
\toprule
\textbf{Method} & \textbf{Elevated Viewpoint} & \textbf{Rotated Viewpoint}  \\ \midrule
BC (finetune) & 2/10  & 3/10  \\
\textbf{\methodname (Ours)} & \textbf{6/10}  & \textbf{7/10}  \\
\bottomrule
\end{tabular}
\vspace{-0.1cm}
\caption{\footnotesize{\textbf{Comparison of \methodname and BC (finetune), when evaluated on novel camera viewpoints} with elevated and rotated cameras as shown in Figure~\ref{fig:multviewpoint} for the croissant task. Observe that \methodname still outperforms BC (finetune) in this setting and attains more than 2x success rate of BC (finetune).}}
\label{tab:viewpoint_comp}
\end{table}

\subsubsection{Expanded Discussion: Why Does \methodname Outperform BC-based methods, Even With Demonstration Data?}

One natural question to ask given the results in this paper is: why does utilizing an offline RL method for pre-training and finetuning as in \methodname outperform BC-based methods even though the dataset is quite ``BC-friendly'', consisting of only demonstrations? One might speculate that an answer to this question is that our BC baseline can be tuned to be much better. However, note that our BC baseline is not suboptimally tuned. We utilize the procedure prescribed by prior work~\citep{ebert2021bridge} for tuning BC as we discuss in Appendix~\ref{app:hyperparams}. In addition, the fact that \textbf{BC (joint)} does actually outperform \textbf{CQL (joint)} in many of our experiments, indicates that our BC baselines are well-tuned. To explain the contrast to \citet{ebert2021bridge}, note that the setup in this prior work utilized many more target task demonstrations ($\geq 50$ demonstrations from the target task) compared to our evaluations, which might explain why our BC-baseline numbers are lower in an absolute sense. Therefore, the technical question still remains: why  would we expect \methodname to perform better than BC? We will attempt to answer this question using some empirical evidence and visualizations. Also, we will aim to provide intuition for why our approach \methodname outperforms the baseline.

\begin{figure}[h]
\centering
\vspace{-0.4cm}
  \includegraphics[width=0.83\linewidth]{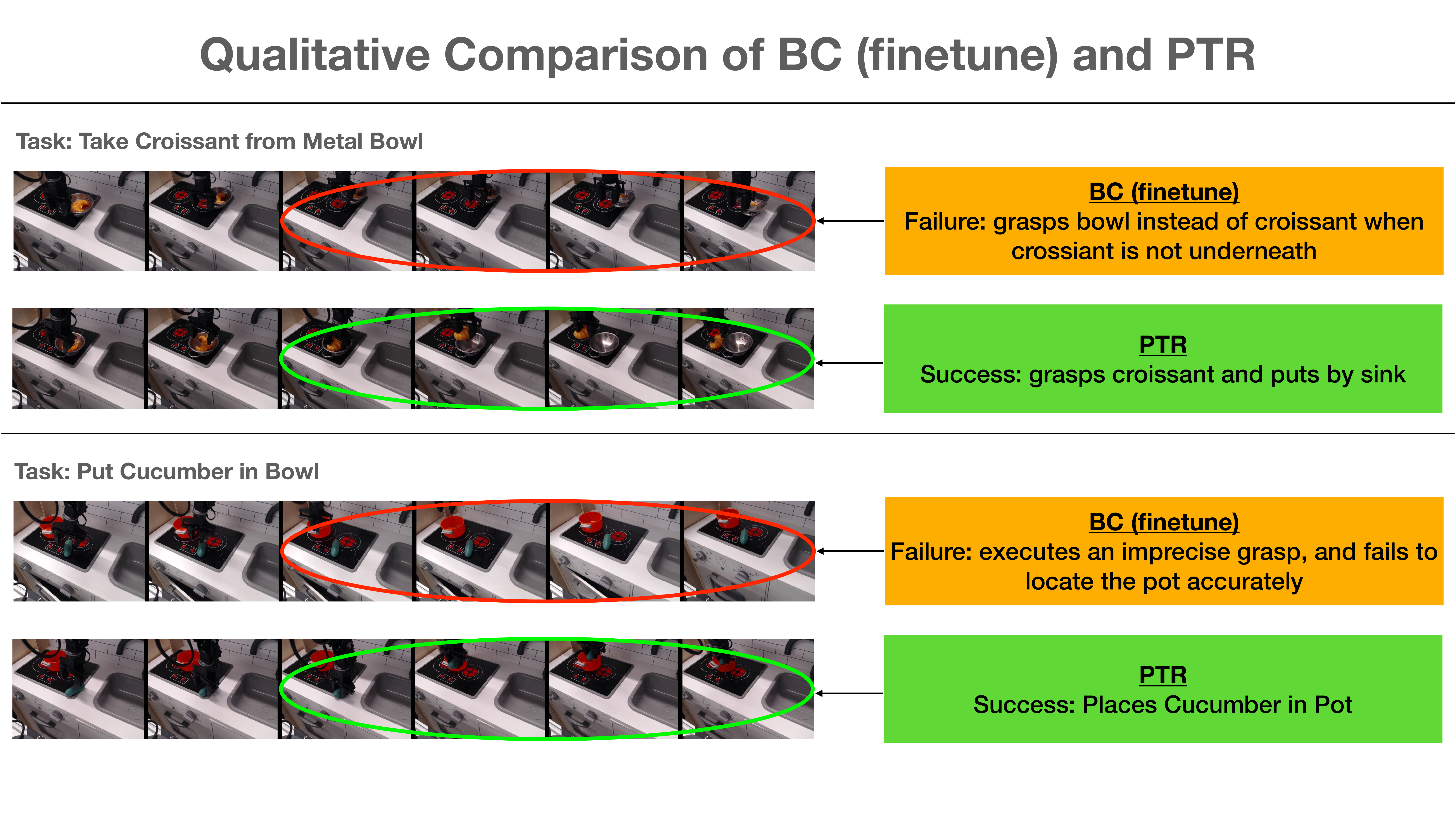}
  \vspace{-0.5cm}
  \caption{\footnotesize \textbf{Qualitative successes of \methodname visualized alongside failures of BC (finetune).} As an example, observe that while \methodname is accurately able to reach to the croissant and grasp it to solve the task, BC (finetune) is imprecise and grasps the bowl instead of the croissant resulting in failure.}
  \label{fig:dumb_behavior2}
  \vspace{-0.3cm}
\end{figure}

\textbf{To begin answering this question,} it is instructive to visualize some failures for a BC-based method and qualitatively attempt to understand why BC is worse than utilizing \methodname. We visualize some evaluation rollouts for \textbf{BC (finetune)} and \methodname as film strips in {Figure~\ref{fig:dumb_behavior2}}. Specifically, we visualize evaluation rollouts that present a challenging initial state. For example, for the rollout from the take croissant out of metallic pot task, the robot must first accurately position itself over the croissant before executing the grasping action. Similarly, for the rollout from the cucumber task, the robot must accurately locate the bowl and precisely try to grasp the cucumber. Observe in {Figure~\ref{fig:dumb_behavior}} that \textbf{BC (finetune)} typically fails to accurately reach the objects of interest (croissant and the bowl) and executes the grasping action prematurely. On the other hand, \methodname is more robust in these situations and is able to accurately reach the object of interest before it executes the grasping action or the releasing action. Why does this happen?  

\textbf{To understand why this happens}, one mental model is to appeal to the critical states argument from \citet{kumar2022should}. Intuitively, this argument suggests that in tasks where the robot must precisely accomplish actions at only a few specific states (called ``\textbf{critical states}'') to succeed, but the actions at other states (called ``non-critical states'') do not matter as much. Thus, offline RL-style methods can outperform BC-based methods even with demonstration data. This is because learning a value function can enable the robot to reason about which states are more important than others, and the resulting policy optimization can ``focus'' on taking correct actions at such critical states. Our real-world evaluation scenarios exhibit such a structure. The majority of the actions that the robot must take to reach the object do not need to be precise as long as they generally move the robot in the right direction. However, in order to succeed, the robot must critically ensure to position the arm is right above the object in a correct orientation and position itself right above the container in which the object must be placed. These are the critical states and special care must be taken to execute the right action in these states. In such scenarios, the argument of \citet{kumar2022should} would suggest that offline RL should be better. We believe that we observe a similar effect in our experiments: the learned BC policies are often not precise-enough at those critical states where taking the right action is critical to success.  

\begin{figure}[h]
\centering
\vspace{-0.4cm}
  \includegraphics[width=0.78\linewidth]{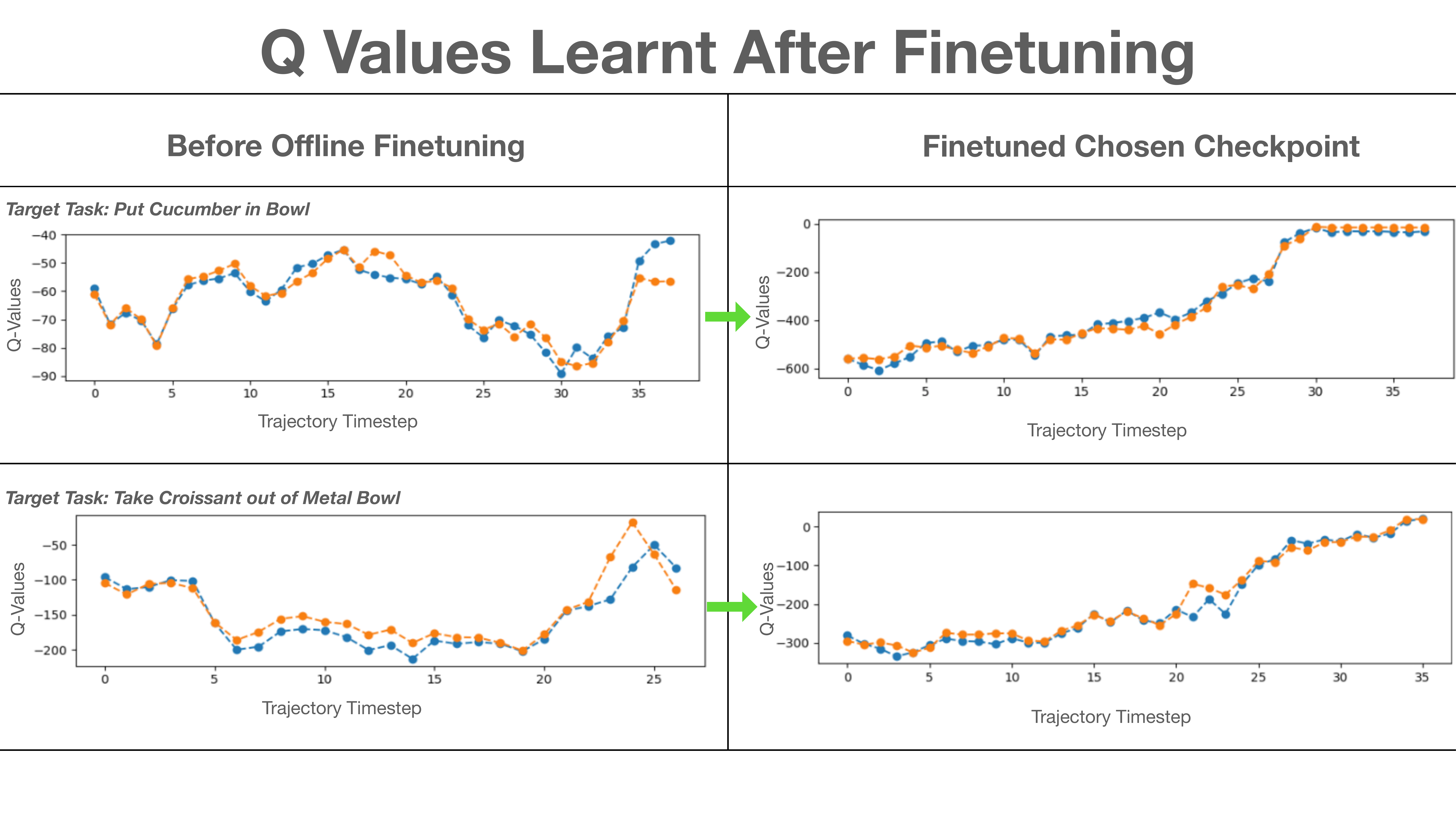}
  \vspace{-0.5cm}
  \caption{\footnotesize \textbf{Evolution of Q-values on the target task} over the process of fine-tuning with \methodname. Observe that while the learned Q-values on \emph{held-out} trajectories from the dataset just at the beginning of Phase 2 (finetuning) do not exhibit a roughly increasing trend, the checkpoint of \methodname we choose to evaluate exhibits a generally increasing trend in the Q-values despite having access to only 10 demonstrations for these target tasks.}
  \label{fig:finetunedQvals2}
  \vspace{-0.3cm}
\end{figure}

{As supporting evidence} to the discussion above, we further visualize the Q-values over held-out trajectories from the target demonstration data that were never seen by \methodname during fine-tuning in {Figure~\ref{fig:finetunedQvals2}}. To demonstrate the contrast, we present the trend in Q-values before fine-tuning and for the checkpoint selected for evaluation after fine-tuning on the target task. Observe that the Q-values for the chosen checkpoint generally increase over the course of the trajectory indicating that the learned Q-function is able to fit well with the target data. Also, the learned Q-function generalizes to held-out trajectories despite the fact that only 10 demonstrations were provided during the fine-tuning phase. This evidence supports the claim that it is reasonable to expect the learned Q-function to be able to focus on the more critical decisions in the trajectory.

\textbf{To further support our hypothesis that \methodname outperforms BC-based methods because the learned value function enables us to learn about ``critical'' decisions}, we run an experiment that essentially runs a weighted version of BC during finetuning, where the weights are provided by exponentiated advantage values, where the advantages are defined as $A_\theta(\bs, \ba) = Q_\theta(\bs, \ba) - \max_{\ba'} Q_\theta(\bs, \ba')$ under a Q-function learned by \methodname. This approaches essentially matches BC finetuning in all aspects: the policy parameterization, the loss function (mean-squared error), and the details of the training are kept identical to our BC baselines, with the exception of an additional weight given by $\exp(A_\theta(\bs, \ba))$ on a given transition $(\bs, \ba, r, \bs')$ observed in the set of limited task-specific demonstrations. We refer to this approach as ``advantage-weighted BC finetuning''.

In contrast to our BC (finetune) results from Table~\ref{tab:scenario4}, where \methodname significantly outperformed BC (finetune), observe in Table~\ref{tab:aw_bc}, that advantage-weighted BC (finetune) performs comparably to \methodname on the two tasks we studied for our analysis. This result is significant since it implies that all other factors kept identical, utilizing the weights given by the Q-function is the crucial factor in improving the performance of BC and avoids the qualitative failure modes associated with BC methods shown in Figure~\ref{fig:dumb_behavior2}.

\begin{table}[h]
\centering
\resizebox{0.99\linewidth}{!}{\begin{tabular}{c|r|r||r}
\toprule
\textbf{Task} & \textbf{BC (finetune)} & \textbf{\methodname (Ours)} &\textbf{Advantage-weighted BC (finetune)}  \\ \midrule
Put cucumber in pot &  0/10 & 5/10 &  {5/10} \\
Take croissant from metal bowl & 3/10 &  7/10  & {6/10} \\
\bottomrule
\end{tabular}}
\vspace{0.1cm}
\caption{\footnotesize{\textbf{Performance of advantage-weighted BC} on two tasks from Table~\ref{tab:scenario4}. Observe that weighting the BC objective using advantage-weights computed using the Q-function learned by \methodname leads to much better performance than standard BC (finetune), and close to PTR. This test indicates that the Q-function in \methodname allows us to focus on more critical points, thereby preventing the failures discussed in Figure~\ref{fig:dumb_behavior2}.}}
\label{tab:aw_bc}
\end{table}

\subsubsection{Hyperparameters for \methodname and Baseline Methods}
\label{app:hyperparams}

In this section, we will present the hyperparameters we use in our experiments and explain how we tune the other hyperparameters for both our method \methodname and the baselines we consider.  

\textbf{\methodname.} Since \methodname utilizes CQL as the base offline RL method, it trains two Q-functions and a separate policy, and maintains a delayed copy of the two Q-functions, commonly referred to as target Q-functions. We utilize completely independent networks to represent each of these five models (2 Q-functions, 2 target Q-functions, and the policy). We also do not share the convolutional encoders among them. As discussed in the main text, we rescaled the action space to $[-1, 1]^{|\mathcal{A}|}$ to match the one used by actor-critic algorithms, and utilized a Tanh squashing function at the end of the policy. We used a CQL $\alpha$ value of 10.0 for our pick-and-place experiments. The rest of the hyperparameters for training the Q-function, the target network updates, and the policy are taken from the standard training for image-based CQL from \citet{singh2020cog} and are presented in Table~\ref{tab:hparams_cql} below for completeness. The hyperparameters we choose are essentially the network design decisions of \textbf{(1)} utilizing group normalization instead of batch normalization, \textbf{(2)} utilizing learned spatial embeddings instead of standard mean pooling, \textbf{(3)} passing in actions at each of the fully connected layers of the Q-network and the hyperparameter $\alpha$ in CQL that must be adjusted since our data consists of demonstrations. We will ablate the new design decisions explicitly in Appendix~\ref{app:design}.

\begin{table*}[h]
\centering
\begin{tabular}{l|c}
\toprule
\textbf{Hyperparameter} & \textbf{Value}\\  \midrule
Q-function learning rate & 3e-4 \\
Policy learning rate & 1e-4 \\
Target update rate & 0.005 (soft update with Polyak averaging) \\
Optimizer type & Adam \\
Discount factor $\gamma$ & 0.96 (since trajectories have a length of only about 30-40) \\
Use terminals & True \\
Reward shift and scale & shift = -1, scale = 10.0 \\
CQL $\alpha$ & 10.0 \\
Use Color Jitter & True \\
Use Random Cropping & True \\
\bottomrule
\end{tabular}
\vspace{0.07cm}
\caption{\footnotesize{\textbf{Main hyperparameters for CQL training in our real-world experiments.} In the simulation, we utilize a smaller $\alpha$ for CQL, $\alpha=1.0$, and a larger discount $\gamma = 0.98$ since trajectories in the simulation are about 60-70 timesteps in length. }}
\label{tab:hparams_cql}
\end{table*}

The only other hyperparameter used by \methodname is the mixing ratio $\tau$ that determines the proportion of samples drawn from the pre-training dataset and the target dataset during the offline finetuning phase in \methodname. We utilize $\tau = 0.7$ for our experiments with \methodname in the main paper, and use $\tau = 0.9$ for the additional experiments we added in the Appendix. This is because $\tau=0.9$ (more bridge data, and a smaller amount of target data) was helpful in scenarios with very limited target data.  

In order to perform checkpoint selection for \methodname, we utilized the trends in the learned Q-values over a set of held-out trajectories on the target data as discussed in Section~\ref{sec:design_choices}. We did not tune any other algorithmic hyperparameters for CQL, as these were taken directly from \citep{singh2020cog}.  

\textbf{BC (finetune).}
We trained BC in a similar manner as \citet{ebert2021bridge}, utilizing the design decisions that this prior work found optimal for their experiments. The policy for BC utilizes the very same ResNet 34 backbone as our RL policy since a backbone based on ResNet 34 was found to be quite effective in \citet{ebert2021bridge}. Following the recommendations of \citet{ebert2021bridge} and based on result trends from our own preliminary experiments, we chose to not utilize the tanh squashing function at the end of the policy for any BC-based method, but trained a deterministic BC policy that was trained to regress to the action in the demonstration with a mean-squared error (MSE) objective. 

\begin{table}[h]
\centering
\begin{tabular}{l|c}
\toprule
\textbf{Hyperparameter} & \textbf{Value}\\  \midrule
Policy learning rate & 1e-4 \\
Optimizer type & Adam \\
Use Color Jitter & True \\
Use Random Cropping & True \\
Dropout & 0.4 \\
\bottomrule
\end{tabular}
\vspace{0.07cm}
\caption{\footnotesize{\textbf{Main hyperparameters for Behavior Cloning Baseline Training in our real-world and simulation experiments.} Note: architecture design choices follow closely to \methodname design choices.}}
\label{tab:hparams_cql}
\vspace{-0.4cm}
\end{table}

In order to perform cross-validation, checkpoint, and model selection for our BC policies, we follow guidelines from prior work~\citep{ebert2021bridge,emmons2021rvs} and track the MSE on a held-out validation dataset similar to standard supervised learning. We found that a ResNet 34 BC policy attained the smallest validation MSEs in general, and for our evaluations, we utilized a checkpoint of a ResNet 34 BC policy that attained the smallest MSE.   

Analogous to the case of \methodname discussed above, we also ablated the performance of BC for a set of varying values of the mixing ratio $\tau$, but found that a large value of $\tau = 0.9$ was the most effective for BC, and hence utilized $\tau = 0.9$ for BC (finetune) and BC (joint).

\textbf{BC (joint) and CQL (joint).} The primary distinction between training \textbf{BC (joint)} and \textbf{BC (finetune)} and correspondingly, \textbf{CQL (joint)} and \methodname was that in the case of joint training, the target dataset was introduced right at the beginning of Phase 1 (pre-training phase), and we mixed the target data with the pre-training data using the same value of the mixing ratio $\tau$ used in for our fine-tuning experiments to ensure a fair comparison.

{
\textbf{Few-shot offline meta-RL (MACAW)~\citep{2020arXiv200806043M}:} We compare to two variants of this algorithm and perform an \textbf{extensive} sweep over several hyperparameters, shown in Table~\ref{tab:hparams_macaw}. 
}

{
We trained two different variants of MACAW in our evaluation: \textbf{(1)} Pre-training on the bridge data in Scenario 3 and then fine-tuning on target data of interest, and \textbf{(2)} adapting a set of existing task identifiers to the target task of interest utilizing the same pre-training and fine-tuning domains. We performed early stopping on the meta-training based on validation losses. From there, we started the meta-testing phase, adapting to the target domain of interest. Following \citet{2020arXiv200806043M}, we use a task mini-batch of 8 tasks at each step of optimization rather than using all of the training tasks. We clipped the advantage weight logits to the scale of 20 and attempted to utilize a policy network with a fixed and learned standard deviation. Additionally, we varied the number of Adaptation steps following prior work. Our evaluation protocol for MACAW entails utilizing the validation losses to choose an initial checkpoint for evaluation. Then, we consider checkpoints in the neighborhood ($\pm$ 50K gradient steps) to for evaluations as well and chose the max over all of these checkpoints as the final evaluation success rate.
}

{
Quantitatively, as seen in Table~\ref{tab:scenario4}, MACAW was unable to get non-zero success rates on any of the tasks we study. However, we did qualitatively observe nontrivial behavior seen in our evaluation rollouts. For instance, we found that the policies trained via MACAW could consistently grasp the object of interest but were unable to localize where to place the object correctly. Several trials involved hovering around with the object of interest and not placing the object in the container. Other trials involved the agent failing to grasp the object.
}

\begin{table}[h]
\centering
\begin{tabular}{l|c}
\toprule
\textbf{Hyperparameter} & \textbf{Value}\\  \midrule
Optimizer & Adam \\
Outer Policy learning rate & 1e-4 \\
Outer Value learning rate & 1e-5, 1e-6 \\
Inner Policy learning rate & 1e-2, 1e-3 \\
Inner Value learning rate & 1e-3, 1e-4 \\
Auxilary Advantage Coefficient & 1e-2, 1e-3, 1e-4 \\
Policy Parameterization & Fixed std, Learned std \\
AWR Policy Temperature & 1, 10, 20 \\
Number of Adaptation Steps & 1, 2, 3 \\
Task Batch Size & 8 \\
Train Adaptation Batch Size & 64 \\
Eval Adaptation Batch Size & 64 \\
Max Advantage Clip & 20 \\
Use Color Jitter & True \\
Use Random Cropping & True \\

\bottomrule
\end{tabular}
\vspace{0.07cm}
\caption{{\footnotesize{\textbf{Main hyperparameters for Training MACAW~\citep{2020arXiv200806043M} in our real-world experiments.} Note: architecture design choices follow closely to \methodname design choices but hyperparameter design choices follow closely the suggestions in \citet{2020arXiv200806043M}.}}}
\label{tab:hparams_macaw}
\vspace{-0.2cm}
\end{table}

\textbf{Pre-trained R3M initialization~\citep{nair2022r3m}:} Next we compare \methodname to utilizing an off-the-shelf pre-trained representation given by R3M~\citep{nair2022r3m}. We compare two baselines that attempt to train an MLP policy on top of the R3M state representation by using BC (finetuning) and CQL (finetuning) respectively. To ensure that this baseline is well-tuned, we tried a variety of network sizes with 2, 3 or 4 MLP layers and also tuned the hidden dimension sizes in [256, 512, 1024]. We also utilized dropout as regularization to prevent overfitting and tuned a variety of values of dropout probability in [0, 0.1, 0.2, 0.4, 0.6, 0.8]. We observe in Table~\ref{tab:scenario4}, that on the four tasks we evaluate on, \methodname outperforms R3M, which indicates that training on the bridge dataset can indeed give rise to effective visual representations that are more suited to finetuning in our setting. The numbers we report in the table are the best over each parametric policy corresponding to each hyperparameter in our ablation. Checkpoint selection was done utilizing early stopping which is the last iteration where the validation error stops decreasing. Learning curves for this baseline can be found in our Anonymous Website.

\textbf{Pre-trained MAE initialization~\citep{he2111masked}:}
We took a similar training procedure to R3M for our MAE representation. We used an MAE trained on every image from the bridge dataset \citet{ebert2021bridge}. We then fine-tuned a specific target task with a similar ablation on network size, hidden dimension size, and regularization techniques such as dropout.  We observe in Table~\ref{tab:rep_learning_comparison}, that on the four tasks we evaluate on, \methodname outperforms R3M, which indicates that training on the bridge dataset can indeed give rise to effective visual representations that are more suited to finetuning in our setting. The numbers we report in the table are the best over each parametric policy corresponding to each hyperparameter in our ablation. Checkpoint selection was done utilizing early stopping which is the last iteration where the validation error stops decreasing. 

\textbf{Policy expressiveness study.}
We considered two policy expressiveness choices for BC to compare with our reference BC implementation that is implemented with a set of MLP layers. The first of the two choices was an \textbf{autoregressive policy} where the 7-dimensional action space was discretized into 100 bins. Each action was then predicted autoregressively conditioned on the observation, task id, and the action component from the previous dimension(s). The second approach was with the BeT Architecture from \citet{shafiullah2022behavior}. We utilized the reference implementation from the paper with the default suggested hyperparameters for this set of ablations. The window size for the MinGPT transformer was ablated over between 1, 2, and 10.

\subsection{Validating the Design Choices from Section~\ref{sec:design_choices} via Ablation Studies}
\label{app:design}

In this section, we will present ablation studies aimed to validate the design choices utilized by \methodname. We found these design choices quite crucial for attaining good performance. The concrete research questions we wish to answer are: \textbf{(1)} How important is utilizing a large network for attaining good performance with \methodname, and how does the performance of \methodname scale with the size of the Q-function?, \textbf{(2)} How effective is a learned spatial embedding compared to other approaches for aggregating spatial information? \textbf{(3)} Is concatenating actions at each fully-connected layer of the Q-function crucial for good performance?, \textbf{(4)} Is group normalization a good alternative to batch normalization? and \textbf{(5)} How does our choice of creating binary rewards for training affect the performance of \methodname?. We will answer these questions next.

\begin{figure}
\includegraphics[width=0.9\linewidth]{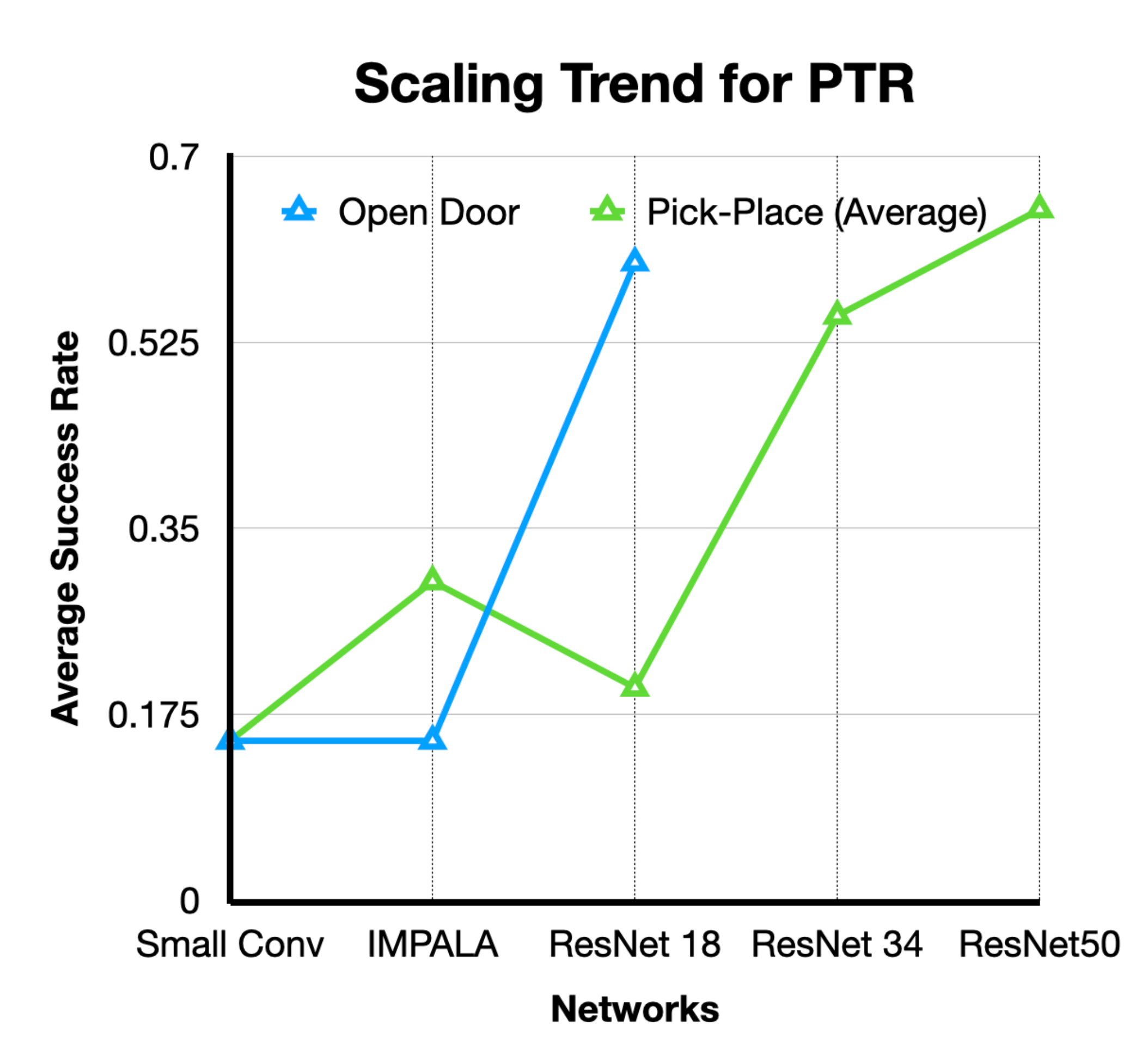}
\caption{\footnotesize{\label{fig:scaling_ptr2} \textbf{Scaling trends for \methodname} on the open door task from Scenario 2, and average over two pick and place tasks (take croissant out of the metallic pot and put cucumber in the bowl) from Scenario 3. Note that more high capacity and expressive function approximators lead to the best results.}}
\end{figure}

\textbf{Highly expressive Q-networks are essential for good performance.} To assess the importance of highly expressive Q-functions, we evaluate the performance of \methodname with varying sizes and architectures on three tasks: the open door task from Scenario 2, and the put cucumber in the pot and take croissant out of metallic bowl tasks from Scenario 3. Our choice of architectures is as follows: \textbf{(a)} a standard three-layer convolutional network typically used by prior work for DM-control tasks (see for example, \citet{kostrikov2021offline}), \textbf{(b)} an IMPALA~\citep{espeholt2018impala} ResNet that consists of 15 convolutional layers spread across a stack of 3 residual blocks, \textbf{(c)} ResNet 18 with group normalization and learned spatial embeddings, \textbf{(d)} ResNet 34 that we use in our experiments, and \textbf{(e)} an even bigger ResNet 50 with group normalization and learned spatial embeddings. 

We present our results in {Figure~\ref{fig:scaling_ptr}}. To obtain more accurate scaling trends, we plot the trend in the average success rates for the pick and place tasks from Scenario 3 along with the trend in the success rate for the open door task separately since these tasks use different pre-training datasets. Observe that the performance of smaller networks (Small, IMPALA) is significantly worse than the ResNet in the door-opening task. For the pick and place tasks that contain a much larger dataset, Small, IMPALA, and ResNet18 all perform much worse than ResNet 34 and ResNet 50. We believe this result is quite exciting since it highlights the possibility of actually benefitting from using highly-expressive neural network models with TD-learning based RL methods trained on lots of diverse multi-task data (contrary to prior work~\citep{lee2022multi}). We believe that this result is a valuable starting point for further scaling and innovation.

\textbf{Learned spatial embeddings are crucial for performance.} Next we study the impact of utilizing the learned spatial embeddings for encoding spatial information when converting the feature maps from the convolutional stack into a vector that is fed into the fully-connected part of the Q-function. We compare our choice to utilizing a spatial softmax as in \citet{ebert2021bridge}, and also global average pooling (GAP) that simply averages over the spatial information, typically utilized in supervised learning with ResNets.

\begin{table}[h]
\centering
\begin{tabular}{l|r}
\toprule
\textbf{Method} & \textbf{Success rate}\\  \midrule
PTR with spatial softmax & 4/10 \\
PTR with global average pooling & 4/10 \\
\midrule
PTR with learned spatial embeddings \textbf{(Ours)} & \textbf{7/10} \\
\bottomrule
\end{tabular}
\vspace{0.1cm}
\caption{\footnotesize{\textbf{Ablation of \methodname with spatial softmax and GAP on the croissant task.} Observe that \methodname with learned spatial embeddings performs significantly better than using a spatial softmax or global average pooling.}}
\label{tab:spatial}
\end{table}

As shown in {Table~\ref{tab:spatial}} learned spatial embeddings outperform both of these prior approaches on the put croissant in pot task. We suspect that spatial softmax does not perform much better than the GAP approach since the softmax operation can easily get saturated when running gradient descent to fit value targets that are not centered in some range, which would effectively hinder its expressivity. This indicates that the approach of retaining spatial information like in \methodname is required for attaining good performance.

\textbf{Concatenating actions at each layer is crucial for performance.} Next, we run \methodname without passing in actions at each fully connected layer of the Q-function on the take croissant out of metallic bowl task and only directly concatenate the actions with the output of the convolutional layers before passing it into the fully-connected component of the network. On the croissant task, we find that not passing in actions at each layer only succeeds in \textbf{2/10} evaluation rollouts, which is significantly worse than the default \methodname which passes in actions at each layer and succeeds in \textbf{7/10} evaluation rollouts (Table~\ref{tab:action_sep}).

\begin{table}[h]
\centering
\begin{tabular}{l|r}
\toprule
\textbf{Method} & \textbf{Success rate}\\  \midrule
PTR without actions passed in at each FC layer & 2/10 \\
PTR with actions passed in at each FC layer (Ours) & \textbf{7/10} \\
\bottomrule
\end{tabular}
\vspace{0.1cm}
\caption{\footnotesize{\textbf{Ablation of \methodname with actions passed in at each layer.} Observe that passing in actions at each fully-connected layer does lead to quite good performance.}}
\label{tab:action_sep}
\end{table}

\textbf{Group normalization is more consistent than batch normalization.} Next, we ablate the usage of group normalization over batch normalization in the ResNet 34 Q-functions that \methodname uses. We found that batch normalization was generally harder to train to attain Q-function plots that exhibit a roughly increasing trend over the course of a trajectory. That said, on some tasks such as the croissant in pot task, we did get a reasonable Q-function, and found that batch normalization can perform well. On the other hand, on the put cucumber in pot task, we found that batch normalization was really ineffective. These results are shown in {Table~\ref{tab:batch_norm}}, and they demonstrate that batch normalization may not be as consistent and reliable with \methodname as group normalization.

\begin{table}
\centering
\scalebox{0.75}{
\begin{tabular}{l|r|r}
\toprule
\textbf{Method} & \textbf{Croissant out of metallic bowl} & \textbf{Cucumber in pot} \\  \midrule
PTR with batch norm. (relative) & + 28.0\% (7/10 $\rightarrow$ 9/10)& - 60.0\% (5/10 $\rightarrow$ 2/10) \\
\bottomrule
\end{tabular}
}
\vspace{0.1cm}
\caption{\footnotesize{\textbf{Relative performance of \methodname with batch normalization with respect to \methodname with group normalization.} Observe that while utilizing batch normalization in \methodname can be sometimes more effective than using group normalization (e.g., take croissant out of metallic bowl task), it may also be highly ineffective and can reduce success rates significantly in other tasks. The performance numbers to the left of the $\rightarrow$ correspond to the performance of \methodname with group normalization and the performance to the right of $\rightarrow$ is the performance with batch normalization.}}
\label{tab:batch_norm}
\end{table}

\textbf{Choice of the reward function.} Finally, we present some results that ablate the choice of the reward function utilized for training \methodname from data that entirely consists of demonstrations. In our main set of experiments, we labeled the last three timesteps of every trajectory with a reward of +1 and annotated all other timesteps with a 0 reward. We tried perhaps the most natural choice of labeling only the last timestep with a 0 reward on the croissant task and found that this choice succeeds \textbf{0/10} times, compared to annotating the last three timesteps with a +1 reward which succeeds \textbf{7/10} times. We suspect that this is because only annotating the last timestep with a +1 reward is not ideal for two reasons: first, the task is often completed in the dataset much earlier than the observation shows the task complete, and hence the last-step annotation procedure induces a non-Markovian reward function, and second, only labeling the last step with a +1 leads to overly conservative Q-functions when used with \methodname, which may not lead to good policies.

\subsection{More Details on Online Fine-tuning}
\label{app:online_finetuning}

\textbf{Offline pre-training.}
For both PTR and BC baseline, we used 40 open-door demonstrations as target task data and combined them with the Bridge Dataset to pre-train the policy. To reduce the training time in the real system, we used ResNet 18 backbones.

\textbf{Reset policy.}
For the reset policy, we additionally collected 22 close-door demonstrations as the target task data and pre-trained the policy with PTR. Similar to the open-door policy, we used ResNet 18 backbones to save training time.

\textbf{Reward classifier.} We used a ResNet 34 classification model and trained it to detect whether the door is open or closed from visual inputs. For the training data, we manipulated the robot to collect around 20 positive and negative trajectories for both open and closed doors.

\textbf{Method.} As shown by \citet{nakamoto2023calql} in simulation, offline value function initializations that learn conservative Q-functions may not be effective at fine-tuning if the learned Q-values are not at the same scale as the ground-truth return of the behavior policy. While this property does not affect offline performance, it is crucial to enforce this property during fine-tuning. That said, this property can be ``baked in'' by simply preventing the CQL regularizer from minimizing the learned Q-function if its values fall below the Monte-Carlo return of the trajectories in the dataset. Therefore, for the online fine-tuning experiment, we incorporate this constraint into PTR.

\textbf{Hyperparameters.} 
For both online fine-tuning with \methodname and SACfD, we performed the experiment by mixing the Bridge Dataset, offline target data, and the online data in a ratio ($\beta$) of 0.35, 0.15, and 0.5. For \methodname, we used the CQL alpha value of 5 for the offline phase and 0.5 for the online phase. 

\textbf{Evaluation.} The results shown in Figure~\ref{fig:online_door} were evaluated autonomously every 5K environment step during the online fine-tuning. Each evaluation was assessed with 10 trials, one from each initial position. The results shown in Figure~\ref{tab:online-finetune} were additionally evaluated over 3 trials from each initial position, using the offline initialization and the final checkpoint obtained after 20K environment steps of online fine-tuning.

\begin{figure}
\includegraphics[width=\linewidth]{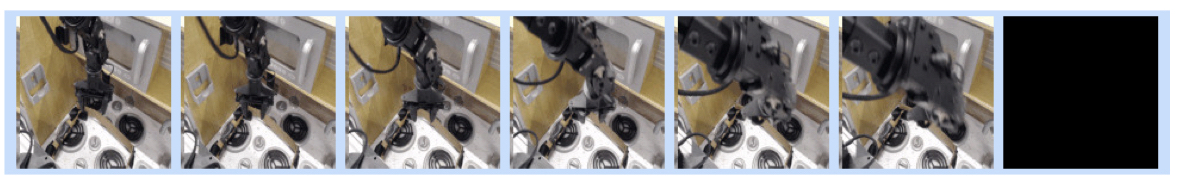}
\caption{\footnotesize{\label{fig:dangerous-actions} \textbf{Example of unsafe behaviors when running SACfD.} The robot collides with the camera during online exploration, resulting in a system crash.}}
\end{figure}

\end{document}

%% file: defs.tex
\newcommand{\policy}{\pi}

\newcommand{\transitions}{T}

\newcommand{\bs}{\mathbf{s}}
\newcommand{\ba}{\mathbf{a}}

\newcommand{\methodname}{PTR\xspace}

%% file: math_commands.tex
\usepackage{amsmath,amsfonts,bm}

\def\eqref#1{equation~\ref{#1}}

\def\1{\bm{1}}

\DeclareMathAlphabet{\mathsfit}{\encodingdefault}{\sfdefault}{m}{sl}
\SetMathAlphabet{\mathsfit}{bold}{\encodingdefault}{\sfdefault}{bx}{n}

\newcommand{\E}{\mathbb{E}}



%% file: introduction.tex
\vspace{0.1cm}
\section{Introduction}
\label{sec:intro}
\vspace{0.1cm}

Robotic learning methods based on reinforcement learning (RL) or imitation learning (IL) have led to impressive results~\citep{levine2016end,kalashnikov2018qtopt, young2020visual, kalashnikov2021mt,ahn2022can}, but the generalization abilities of policies learned this way are typically limited by the quantity and breadth of training data available. In practice, the cost of real-world data collection for each task means that such methods often use smaller datasets, which leads to more limited generalization. A natural way to circumvent this limitation is to incorporate existing diverse robotic datasets into the training pipeline of a robot learning algorithm, analogously to how pre-training on diverse prior datasets has enabled rapid fine-tuning in supervised learning. How can we devise methods that enable effective pre-training for robotic RL?

In most cases, answering this question requires devising a method that can pre-train on existing data from a wide range of tasks and domains, and then provide a good starting point for efficiently learning a \emph{new} task in a \emph{new} domain. Prior approaches utilize such existing data by running imitation learning (IL)~\citep{young2020visual,ebert2021bridge,shafiullah2022behavior} or by using representation learning~\citep{nair2022r3m} methods for pre-training and then fine-tuning with imitation learning. However, this may not necessarily lead to representations that can reason about the consequences of their actions. In contrast, end-to-end RL can offer a more general paradigm, that can be effective for both pre-training and fine-tuning, and is applicable even when assumptions in prior work are violated. Hence we ask, can we devise a simple and unified framework where \emph{both} the pre-training and fine-tuning process uses RL? This presents significant challenges pertaining to leveraging large amounts of offline multi-task datasets, which would require high capacity models and this can be very challenging \citep{bjorck2021towards}.

In this paper, we show that multi-task offline RL pre-training on diverse multi-task demonstration data followed by offline RL fine-tuning on a very small number of trajectories (as few as 10 trials, maximum 15) or online fine-tuning on autonomously collected data, can indeed be made into an effective robotic learning strategy that can significantly outperform methods based on imitation learning as well as RL-based methods that do not employ pre-training. This is surprising and significant, since prior work~\citep{mandlekar2021what} has suggested that IL methods are superior to offline RL when provided with human demonstrations. Our framework, which we call \methodname (pre-training for robots), is based on the CQL algorithm~\citep{kumar2020conservative}, but introduces a number of design decisions, that we show are critical for good performance and enable large-scale pre-training. These choices include a specific choice of architecture for providing high capacity while preserving spatial information, the use of group normalization, and an approach for feeding actions into the model that ensures that actions are used properly for value prediction. We experimentally validate these design decisions and show that PTR benefits from increasing the network capacity, even with large ResNet-50 architectures, which have never been previously shown to work with offline RL. Our experiments utilize the Bridge Dataset~\citep{ebert2021bridge}, which is an extensive dataset consisting of thousands of trials for a very large number of robotic manipulation tasks in multiple environments. A schematic of PTR is shown in Figure~\ref{fig:system_overview}. 

The main contribution of this work is a demonstration that PTR can enable offline RL pre-training on diverse real-world robotic data, and that these pre-trained policies can be fine-tuned to learn new tasks with just 10-15 demonstrations or with autonomously collected online interaction data in the real world. 
This is a significant improvement over prior RL-based pre-training and fine-tuning methods, which typically require thousands of trials~\citep{singh2020cog,kalashnikov2021mt,julian2020never,chebotar2021actionable,lee2022spend}. We present a detailed analysis of the design decisions that enable offline RL to provide an effective pre-training framework, and show empirically that these design decisions are crucial for good performance. {Although these decisions are based on prior work, we show that the novel combination of these components in PTR is important to make offline RL into a viable pre-training tool that can outperform other approaches.}

%% file: related.tex
\vspace{0.1cm}
\section{Related Work}
\label{sec:related}
\vspace{0.05cm}

A number of prior works have proposed algorithms for offline RL~\citep{fujimoto2018off,kumar2019stabilizing,kumar2020conservative,kostrikov2021offline,kostrikov2021iql,wu2019behavior,jaques2019way,fujimoto2021minimalist,siegel2020keep}. In particular, many prior works study offline RL with multi-task data and devise techniques that perform parameter sharing\citep{wilson2007multi, parisotto2015actor, teh2017distral, espeholt2018impala, hessel2019multi}, or perform data sharing or relabeling~\citep{yu2021conservative,andrychowicz2017hindsight,yu2022leverage,kalashnikov2021mt,xie2021lifelong}. In this paper, our goal is not to develop new offline RL algorithms, but to show that these offline RL algorithms can be an effective tool to pre-train from prior data and then fine-tune on new tasks. We show that a few simple but important design decisions are essential for making offline RL pre-training scalable, and provide detailed experiments on fine-tuning these pre-trained models to new tasks.

Going beyond methods that only perform fine-tuning from a learned initialization with online interaction~\citep{nair2020accelerating,kostrikov2021iql,lee2022offline}, we consider two independent fine-tuning settings: (1) the setting where we do not use any online interaction and fine-tune the pre-trained policy entirely offline, (2) the setting where a limited amount of online interaction is allowed to autonomously acquire the skills to solve the task from a challenging initial condition.
This resembles the problem setting considered by offline meta-RL methods~\citep{li2019multi, dorfman2020offline, mitchell2021offline, pong2021offline,lin2022model}. {However, our approach is simpler as we fine-tune the very same offline RL algorithm that we use for pre-training. In our experiments, we observe that our method, PTR, outperforms the meta-RL method of \citet{mitchell2021offline}. 

Some other prior approaches that attempt to leverage large, diverse datasets via representation learning~\citep{mandlekar2020iris,yang2021representation,yang2021trail,nair2022r3m,he2111masked,xiao2022masked,ma2022vip},
as well as other methods for learning from human demonstrations, such as behavioral cloning methods with expressive policy architectures~\citep{shafiullah2022behavior}.
We compare to some of these methods~\citep{xiao2022masked,nair2022r3m} in our experiments and find that PTR outperforms these methods. We also perform an empirical study to identify the design decisions behind the improved performance of RL-based PTR on demonstration data compared to BC, and find that the gains largely come from the ability of the value function in identifying the most ``critical'' decisions in a trajectory. While some prior works~\citep{mandlekar2021what} shows results that suggest that offline RL underperforms imitation learning when provided with human demonstration data, our results show that offline RL can perform better than BC even with demonstrations, supporting the analysis in \citet{kumar2022should}.

The most closely related to our work are prior methods that run model-free offline RL on diverse real-world data and then fine-tune on new tasks~\citep{singh2020cog,kalashnikov2021mt,julian2020never,chebotar2021actionable,lee2022spend}.
These prior methods typically only consider the setting of \emph{online} fine-tuning, whereas in our experiments, we demonstrate the efficacy of PTR for offline fine-tuning (where we must acquire a good policy for the downstream task using 10-15 demonstrations) \emph{as well as} online fine-tuning considered in these prior works, where we must acquire a new task entirely via autonomous interaction in the real world. 

%% file: prelims.tex
\vspace{0.1cm}
\section{Preliminaries and Problem Statement}
\vspace{0.1cm}
An RL algorithm aims to learn a policy in a Markov decision process (MDP), which is a tuple $\mathcal{M} = (\mathcal{S}, \mathcal{A}, \transitions, r, \mu_0, \gamma)$, where $\mathcal{S}, \mathcal{A}$ denote the state and action spaces, and
$\transitions(\bs' | \bs, \ba)$, $r(\bs,\ba)$ represent the dynamics and reward function respectively. 
$\mu_0(\bs)$ denotes the initial state distribution, and $\gamma \in (0,1)$ denotes the discount factor. The policy $\pi(\ba|\bs)$ learned by RL agents must optimize the long-term cumulative reward, $\max_\policy J(\policy) := \E_{(\bs_t, \ba_t) \sim \pi}[\sum_{t} \gamma^t r(\bs_t, \ba_t)].$ 

\textbf{Problem statement.} Our goal is to learn general-purpose initializations from a broad, multi-task offline dataset and then fine-tune these initializations to specific downstream tasks. We denote the general-purpose offline dataset by $\mathcal{D}$, which is partitioned into $k$ chunks. Each chunk contains data for a given robotic task (e.g., picking and placing a given object) collected in a given domain (e.g., a particular kitchen). See \autoref{fig:system_overview} for an illustration. 
Denoting the task/domain abstractly using an identifier $i$, the dataset can be formally represented as $\mathcal{D} = \cup_{i=1}^k \left(i, \mathcal{D}_i \right)$, where we denote the set of training tasks concisely as $\mathcal{T}_{\text{train}} = [k]$. Chunk $\mathcal{D}_i$ consists of data for a given task identifier $i$, and consists of a collection of transition tuples, $\mathcal{D}_i = \{(\bs^i_j, \ba^i_j, r^i_j, \bs'^i_j)\}_{j=1}^n$ collected by a demonstrator on task $i$. Each task has a different reward function. Our goal is to utilize this multi-task dataset to help train a policy for one or multiple target tasks (denoted without loss of generality as task $\mathcal{T}_{\text{target}} = \{k+1, \cdots, n\}$). 

While the diverse prior dataset $\mathcal{D}$ does not contain any experience for the target tasks, in the offline fine-tuning setting, we are provided with a very small dataset of demonstrations $\mathcal{D}^* := \{\mathcal{D}_{k+1}^*, \mathcal{D}^*_{k+1}, \cdots, \mathcal{D}^*_n\}$ corresponding to each of the target tasks. In our experiments, we use only 10 to 15 demonstrations for each target task, making it impossible to learn the target task from this data alone, such that a method that effectively maximizes performance for the target tasks $\mathcal{T}_\text{target}$ must leverage the prior data $\mathcal{D}$. We also study the setting where we aim to quickly fine-tune the policy learned via offline pre-training and offline fine-tuning using limited amounts of autonomously collected data via online real-world interaction. More details about this setup are provided in Section~\ref{sec:experiments_online}.
 
\textbf{Background and preliminaries.} The Q-value of a given state-action tuple $Q^\pi(\bs, \ba)$ for a policy $\pi$ is the long-term discounted reward attained by executing action $\ba$ at state $\bs$ and following policy $\pi$ thereafter. The Q-function satisfies the Bellman equation $Q^\pi(\bs, \ba) = r(\bs, \ba) + \gamma \mathbb{E}_{\bs', \ba'}[Q^\pi(\bs', \ba')]$. Typical model-free offline RL methods~\citep{fujimoto2018off,kumar2019stabilizing,kumar2020conservative} alternate between estimating the Q-function of a fixed policy $\pi$ using the offline dataset $\mathcal{D}$ and then improving the policy $\pi$ to maximize the learned Q-function. Our system, \methodname, utilizes one such model-free offline-RL method, conservative Q-learning (CQL)~\citep{kumar2020conservative}. We discuss how we adapt CQL for pre-training on diverse data followed by single-task fine-tuning in Section~\ref{sec:method}.

\textbf{Tasks and domains}. We use the Bridge Dataset~\cite{ebert2021bridge} as the source of our pre-training tasks, which we augment with a few additional tasks as discussed in Section~\ref{sec:result}. Our terminology for ``task'' and ``domain'' follows \citet{ebert2021bridge}: a task is a skill-object pair, such as ``put potato in pot'' and a domain corresponds to an environment, which in the case of the Bridge Dataset consists of different toy kitchens, potentially with different viewpoints and robot placements. We assume the new tasks and environments come from the same training distribution, but are not seen in the prior data.

%% file: method.tex
\vspace{0.1cm}
\section{Learning Policies for New Tasks\\ from Offline RL Pre-training}
\label{sec:method}
\vspace{0.1cm}

To effectively solve new tasks from diverse offline datasets, a robotic learning framework must: \textbf{(1)} extract useful skills out of the diverse robotic dataset, and \textbf{(2)} rapidly specialize the learned skills towards an unseen target task, given only a minimal amount of experience from this target task in the form of demonstrations, or collected autonomously by interaction. In this section, we present our framework, \methodname, that provides these benefits by training a single, highly expressive deep network via offline RL, and then specializes it on the target task with a small amount of data. We will first present the key components of our robotic framework in Section~\ref{sec:algorithm} and then discuss our novel technical contributions, the practical design choices that are crucial, in Section~\ref{sec:design_choices}.    

\vspace{0.05cm}
\subsection{The Components of \methodname}
\label{sec:algorithm}
\vspace{0.1cm}

To satisfy both requirements \textbf{(1)} and \textbf{(2)} from above, our framework uses a multi-task offline RL approach, where the policy and Q-function are conditioned on a task identifier. This allows us to share a single set of weights for all possible tasks in the diverse offline dataset, providing a general-purpose pre-training procedure that can use diverse data. 
Once a policy is obtained via this multi-task pre-training process, we adapt this policy for solving a new target task by utilizing a very small amount of target task data or autonomously collected data. We describe the two phases, pre-training and fine-tuning, below:

\textbf{\textbf{Phase 1:} Multi-task offline RL pre-training.} In the first phase, \methodname learns a single Q-function and policy for all tasks $i \in \mathcal{T}_\text{train}$ conditioned on the task identifier $i$, i.e., $Q_\phi(\bs, \ba; i)$ and $\pi_\theta(\ba|\bs, i)$, via multi-task offline RL. We use a one-hot task identifier that imposes minimal assumptions on the task structure. For multi-task offline RL, we use the conservative Q-learning (CQL)~\citep{kumar2020conservative} algorithm, extending it to the multi-task setting. This amounts to training the multi-task Q-function against a temporal difference error objective along with a regularizer that explicitly minimizes the expected Q-value under the learned policy $\pi_\theta(\ba|\bs; i)$,
to prevent overestimation of Q-values for unseen actions, which can lead to poor offline RL performance~\citep{kumar2019stabilizing}. 
Formally, the training objective for our multi-task Q-function, as prescribed by CQL, is given by:
\begin{align*}
\label{eqn:cql_training}
\!\!\!\!\!\!\!\! \min_{\phi}~~ & \alpha\left(\mathop{\mathbb{E}}_{\substack{i \sim \mathcal{T}_\text{train},\\ \bs \sim \mathcal{D}_i, \ba \sim \pi}} \!\!\!\!\left[Q_\phi(\bs,\ba; i)\right] - \!\!\!\mathop{\mathbb{E}}_{\substack{i \sim \mathcal{T}_\text{train},\\ \bs, \ba \sim \mathcal{D}}}\!\!\!\left[Q_\phi(\bs,\ba; i)\right]\right) \\ 
& + \frac{1}{2} \mathop{\mathbb{E}}_{\substack{i \sim \mathcal{T}_\text{train},\\ \bs, \ba, \bs' \sim \mathcal{D}\\\ba' \sim \pi}}\left[\left(Q_\phi(\bs, \ba; i) - r - \gamma Q_{\bar{\phi}}(\bs', \ba')\right)^2 \right].
\end{align*}  
${Q}_{\bar{\phi}}$ denotes the target Q-network, which is a delayed copy of the current Q-network. We train $\phi$ by running gradient descent on the above objective, and then optimize the learned policy to maximize the learned Q-values, along with an additional entropy regularizer as shown below:
\begin{align*}
    \max_{\theta}~~~~ \mathbb{E}_{{i \sim \mathcal{T}_\text{train}, \bs \sim \mathcal{D}_i}}\left[ \mathbb{E}_{\ba \sim \pi_\theta(\cdot|\bs; i)}[Q_\phi(\bs, \ba; i)]  \right] + \beta \mathcal{H}(\pi_\theta).
\end{align*}
At the end of this multi-task offline training phase, we obtain a policy $\pi^\text{off}_\theta$ and Q-function $Q^\text{off}_\phi$, that are ready to be fine-tuned to a new downstream task.

\textbf{Phase 2: Offline or online fine-tuning of $\pi^\text{off}_\theta$ and $Q^\text{off}_\phi$ to a target task $\mathcal{T}_\text{target}$.} In the second phase, PTR attempts to learn a policy to solve one or more downstream tasks by adapting $\pi^\text{off}_\theta$, using a limited set of user-provided demonstrations that we denote $\mathcal{D}^*$, or using a combination of target demonstration data and autonomously collected online data. Our method for the offline fine-tuning setting is simple yet effective: we incorporate the new target task data into the replay buffer of the very same offline multi-task CQL algorithm from the previous phase and resume training from Phase 1. However, na\"ively incorporating the target task data into the replay buffer might still not be effective since this scheme would hardly ever train on the target task data during adaptation due to the large imbalance between the sizes of the few target demonstrations and the large pre-training dataset. 
To address this imbalance, each minibatch passed to multi-task CQL during offline fine-tuning consists of a $\tau$ fraction of transitions from bridge demonstration data and $1 - \tau$ fraction of transitions from the target dataset. By setting $\tau$ to be small, we are able to prioritize multi-task CQL to look at target task data frequently, enabling it to make progress on the downstream task without overfitting.

For the autonomous online fine-tuning setting, we utilize a similar technique and have each mini-batch consist of $\beta$ fraction of transitions from the bridge data and the target demonstration data, and $1 - \beta$ fraction of transitions from the newly collected online data. We alternate between collecting one trajectory and making 10 gradient steps for every single transition collected in the environment. Utilizing a high update to the data ratio allowed us to efficiently train the agent on newly collected online samples from rollouts.

\textbf{Handling task identifiers for new tasks.} The description of our system so far has assumed that the downstream test tasks are identified via a task identifier. In practice, we utilize a one-hot vector to indicate the index of a task. While such a scheme is simple to implement, it is not quite obvious how we should incorporate new tasks with one-hot task identifiers. In our experiments, we use two approaches for solving this problem: first, we can utilize a larger one-hot encoding that incorporates tasks in both $\mathcal{T}_\text{train}$ and $\mathcal{T}_\text{target}$, but not use the indices for $\mathcal{T}_\text{target}$ during pre-training. The Q-function and the policy are trained on these \emph{placeholder} task identifiers only during fine-tuning in Phase 2. 
Another approach for handling new tasks is to not use unique task identifiers for every new task, but rather ``\emph{re-target}'' or re-purpose existing task identifiers for new target tasks in the fine-tuning phase. \methodname provides this option: we can simply assign an already existing task identifier to the target demonstration data before fine-tuning the learned Q-function and the policy. For example, in our experiments in Section~\ref{sec:result} we re-target the put sushi in pot task which uses orange transparent pots to instead put the sushi into a metal pot, which was never seen during training.

A complete overview of our approach is shown in \autoref{fig:system_overview}. We use a value of $\alpha=10.0$ in multi-task CQL and $\tau=0.8$ for mixing the pre-training dataset and the target task dataset in most of our experiments in the real-world, without requiring any domain-specific tuning. For online fine-tuning, we utilized $\alpha=0.5$ to evenly mix between the online and offline datasets. 

\vspace{0.07cm}
\subsection{Important Design Choices and Practical Considerations}
\label{sec:design_choices}
\vspace{0.07cm}
Even though the components discussed in Section~\ref{sec:algorithm} are sufficient to give rise to an offline pre-training and fine-tuning approach, as we show in Section~\ref{fig:experiments}, this approach does not lead very good results on its own. Instead, we must make some crucial design decisions, including designing  neural network architectures that can learn from diverse data with offline RL,
cross-validation metrics to identify policies we expect to be effective after fine-tuning, and the design of the reward functions that can be used to label the pre-training dataset. {We show that making the right choices for these components leads to significant improvement (more than \textbf{3.5x} in final real-world performance; see Appendix~\ref{app:design}). Thus, describing, analyzing, and evaluating these choices is a crucial part of this work that we hope will facilitate applications of offline RL pre-training.}

\begin{figure}
\vspace{-0.5cm}
    \centering
    \setcounter{figure}{1}
  \includegraphics[width=0.85\linewidth]{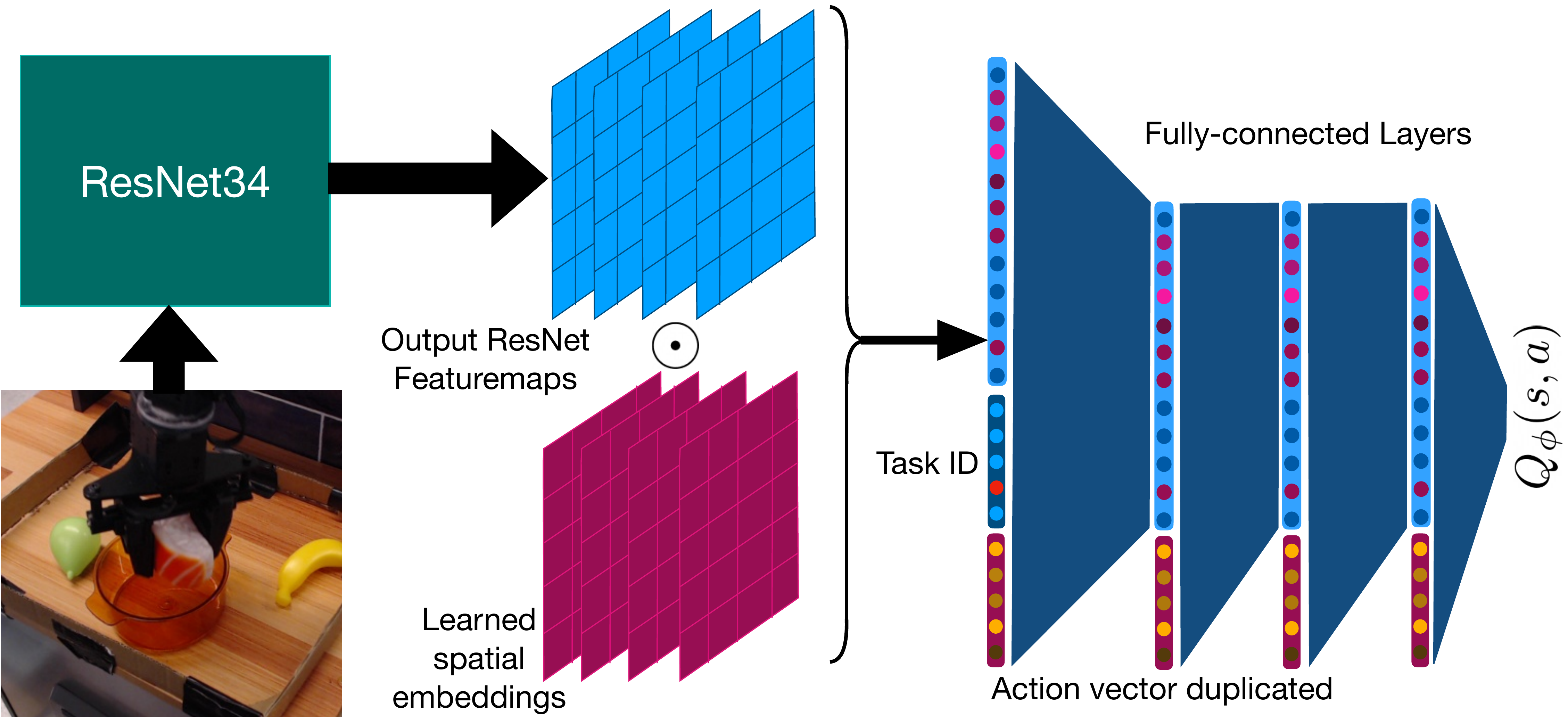}
  \caption{\footnotesize{\textbf{Q-function architecture for \methodname.} The encoder is a ResNet34 with group normalization along with learned spatial embeddings (left). The decoder (right) is a MLP with the action vector duplicated and passed in at each layer. A one-hot task identifier is also passed into the input of the decoder.}}
  \vspace{-0.8cm}
  \label{fig:arch}
\end{figure}
\textbf{Policy and Q-function architectures.} Perhaps the most crucial design decision for our approach is the neural network architecture for representing $\pi^\text{off}$ and $Q^\text{off}$. Since we wish to fine-tune the policy for different tasks, we must use high-capacity neural network models for representing the policy and the Q-function. We experimented with a variety of standard (high-capacity) architectures for vision-based robotic RL. This includes standard convolutional architectures~\citep{singh2020cog} and IMPALA architectures~\citep{espeholt2018impala}. However, we observed in Figure~\ref{fig:scaling_ptr} that these {standard models were unable to effectively handle the diversity of the pre-training data} and performed poorly.
Then, we attempted to utilize standard ResNets~\citep{resnet} (ResNet-18, Resnet-34, and their adaptations to imitation problems from \citet{ebert2021bridge}) to represent $Q_\phi$, but faced divergence challenges similar to prior efforts that use batch normalization~\citep{bjorck2021towards,2019arXiv190205605B} in the Q-network. Batch normalization layers are known to be hard to train with TD-learning~\citep{2019arXiv190205605B} and, therefore, by replacing batch normalization layers with \textcolor{brown}{\textbf{group normalization}} layers~\citep{wu2018group}, we were able to address such divergence issues.
See Appendix~\ref{app:design} for quantitative studies comparing these choices. Unlike prior work~\citep{lee2022multi}, we observed that with group normalization, we attain favorable scaling properties of \methodname: the more the parameters, the better the performance as shown in Figure~\ref{fig:scaling_ptr}.
We also observed that choosing an appropriate method for converting the three-dimensional feature-map tensor produced by the ResNet into a one-dimensional embedding plays a crucial role for learning accurate Q-functions and obtaining functioning policies. Unlike standard ResNet architectures for supervised learning, simply utilizing global average pooling (as used in many classification architectures) performs poorly. Instead we point-wise multiply the learned feature-map with a 3-dimensional parameter tensor before computing sums over the spatial dimensions which allows the network to explicitly encode spatial information. {We refer to this technique as ``\textcolor{brown}{\textbf{learned spatial embeddings}}''}. An illustration of this architecture is provided in \autoref{fig:arch}. As detailed in Appendix~\ref{app:design}, Table~\ref{tab:spatial}, {we find that utilizing this technique leads to improved performance}.

Next, we found that a Q-function $Q_\phi(\bs, \ba)$ obtained by running na\"ive multi-task CQL on the demonstration data tends to not use the action input $\ba$ effectively, due to strong correlations between $\bs$ and $\ba$ in the data, which is almost always the case for narrow, human demonstrations. As a result, policy improvement against such a Q-function overfits to these correlations, producing poor policies. To resolve this issue, we modified the architecture of Q-network to \textcolor{brown}{\textbf{pass the action \textbf{\textit{$\ba$}} as input to every fully-connected layer}} which, as shown in \autoref{fig:arch} and {Appendix~\ref{app:design}}, {Table~\ref{tab:action_sep}}), greatly alleviates the issue {and significantly improves over na\"ive CQL}.

\begin{figure}[h]
\centering
  \includegraphics[width=0.97\linewidth]{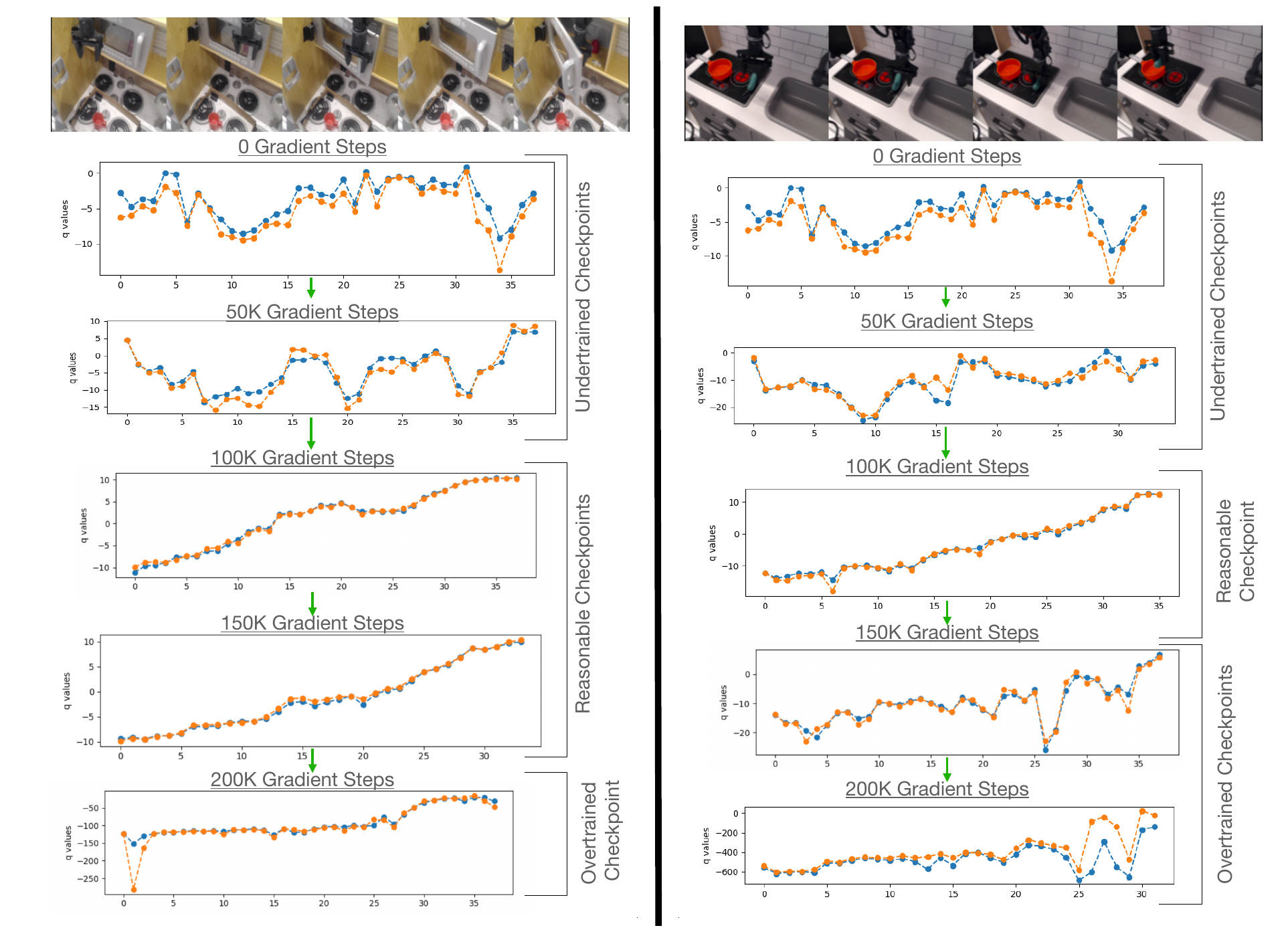}
  \vspace{-0.1cm}
  \caption{\footnotesize \textbf{Evolution of Q-values on the target task over the process of fine-tuning with \methodname.} Observe that while the learned Q-values on \emph{held-out} trajectories from the dataset just at the beginning of Phase 2 (fine-tuning) do not exhibit a roughly increasing trend, we choose to evaluate those checkpoints of \methodname that exhibit a visible more increasing trend in the Q-values despite having access to only 10 demonstrations for these target tasks.}
  \label{fig:moreexreb}
  \vspace{-0.1cm}
\end{figure}

\textbf{Cross-validation during offline fine-tuning.} As we wish to learn task-specific policies that do not overfit to small amounts of data, we must apply the right number of gradient steps during fine-tuning: too few gradient steps will produce policies that do not succeed at the target tasks, while too many gradient steps will give policies that have likely lose the generalization ability of the pre-trained policy. {To handle this trade-off, we adopt the following heuristic as a loose guideline:} we run fine-tuning for many iterations while also plotting the learned Q-values over a held-out dataset of trajectories from the target task as seen in Figure~\ref{fig:moreexreb}. Then for evaluation, we pick the checkpoints that presented a Q-function with the Q-values appearing closest to having a monotonically increasing trend in a trajectory. This is a \emph{relative} guideline and must be performed within the checkpoints observed within a run. The reason for this heuristic choice is that a valid Q-function must be a valid estimator for discounted return, and hence, it must increase over time-steps of a trajectory for a given task.  Of course, this heuristic does not hold for arbitrary sub-optimal offline data, but all of our data comes from human-collected demonstrations. In principle, this heuristic can be wrapped into a metric quantifying degree of monotonicity of the Q-value curve in Figure~\ref{fig:moreexreb}, but in our experiments, we felt this was not necessary: as we show below, we were able to narrow down the checkpoints to essentially one or at most, two checkpoints by just visual inspection. Of course, designing an accurate metric would be helpful for future work. We present two worked-out examples of our checkpoint selection strategy for two tasks from Scenario 1 and Scenario 3 in Figure~\ref{fig:moreexreb}. Observe that checkpoints early in training exhibit Q-values that fluctuate arbitrarily at the beginning of training, which is clearly non-monotonic. This is because of the lack of sufficient gradient steps for fine-tuning the target task. Once sufficient gradient steps are performed, the Q-values visibly improve on the monotonicity property. Training further leads to much flatter Q-values, that are visibly less monotonic.

\begin{figure}[h]
\centering
\vspace{-0.2cm}
  \includegraphics[width=0.65\linewidth]{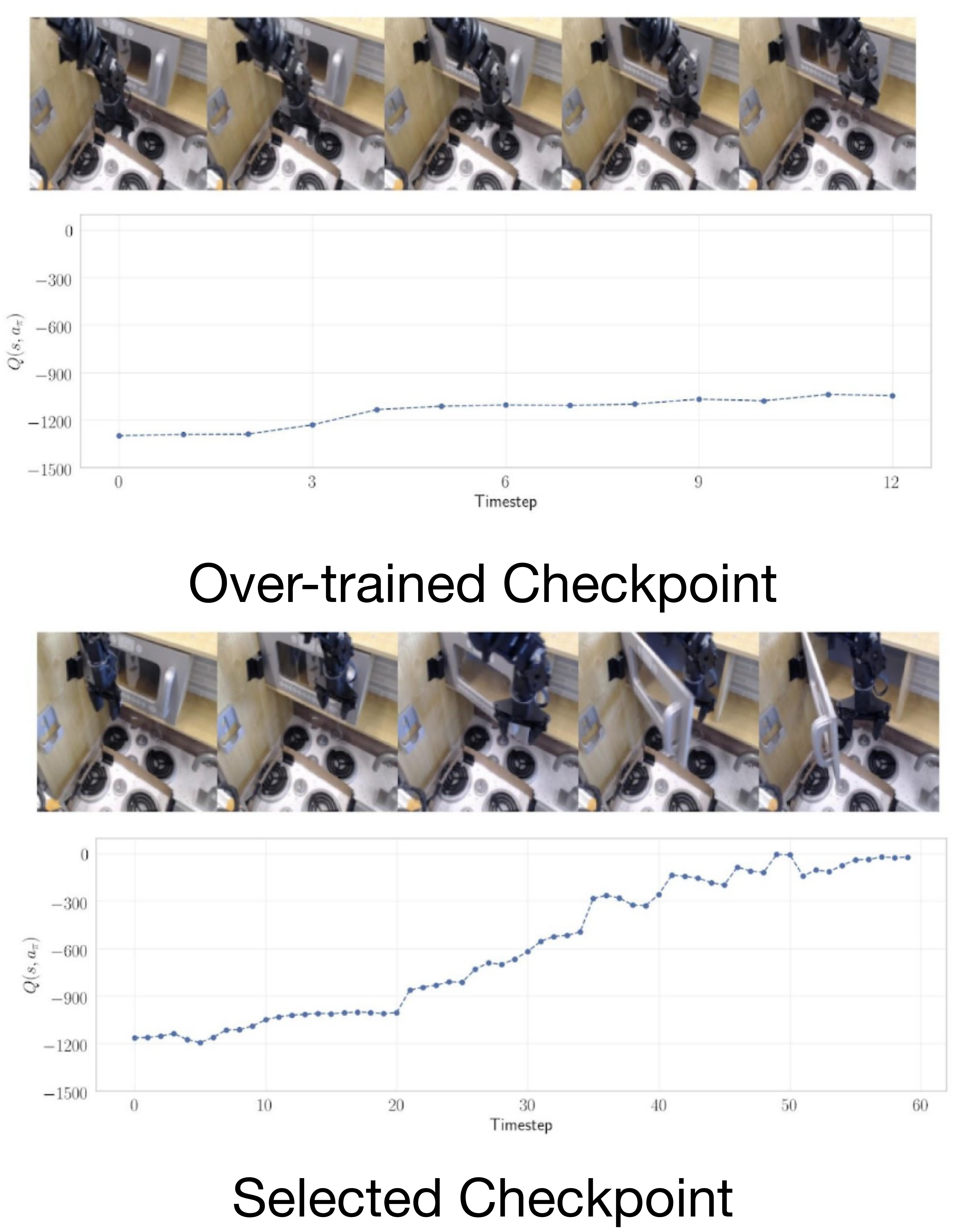}
  \vspace{-0.1cm}
  \caption{\footnotesize \textbf{Performance evaluation of a selected and over-trained checkpoint of \methodname.} We validate our checkpoint selection mechanism on the door opening task. An over-trained checkpoint with nearly flat Q-values fails to solve the task, whereas a checkpoint with visibly increasing Q-values solves the task.}
  \label{fig:validation_door}
  \vspace{-0.6cm}
\end{figure}

To validate our checkpoint selection mechanism, in Figure~\ref{fig:validation_door} we present a film-strip of a sample evaluation of a good and a poor checkpoint as identified by the cross-validation strategy mentioned above. We observe that the checkpoint with more flat Q-values fails to solve the door opening task, whereas the one with a visibly increasing Q-value trend solves the task.

\textbf{Reward specification.} In this paper, we aim to pre-train on existing robotic datasets, such as the Bridge Dataset~\citep{ebert2021bridge}, which consists of human-teleoperated demonstration data. Although the demonstrations are all successful, they are not annotated with any reward function. Perhaps an obvious choice is to label the last transition of each trajectory as success, and give it a +1 binary reward. However, in several of the datasets we use, there can be a 0.5-1.0 second lag between task completion and when the episode is terminated by the data collection. To ensure that a successful transition is not incorrectly labeled as $0$, we utilized the practical heuristic of annotating the last $n=3$ transitions of every trajectory with a reward of $+1$ and and annotated other states with a $0$ reward. We show in Appendix~\ref{app:exp_results} that this provided the best results. 
In principle, more complicated methods of reward labeling~\citep{eysenbach2021replacing} could be used. However, we found the presented rule to be simple and yet effective to learn good policies.

%% file: experiments.tex
\begin{figure*}
\centering
  \includegraphics[width=0.8\linewidth]{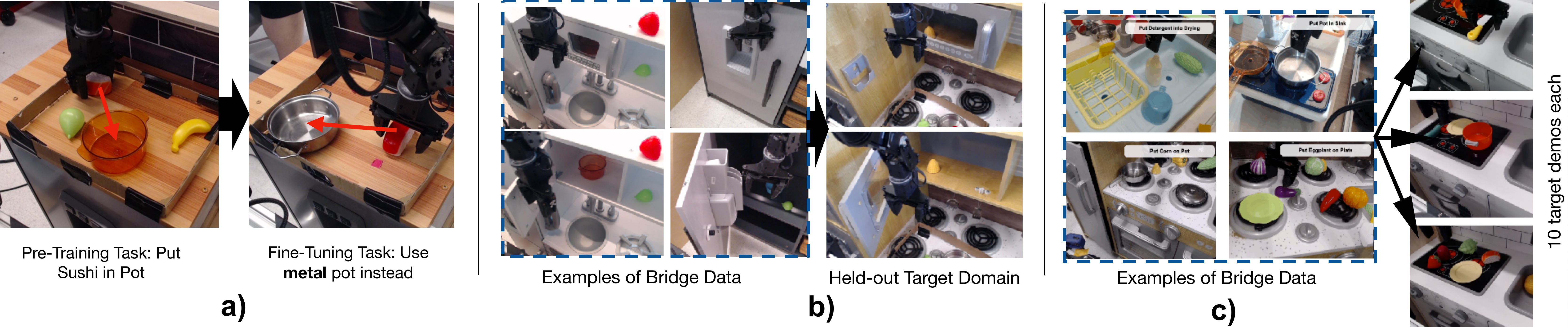}
  \caption{\footnotesize{\textbf{Illustrations of the three real-world experimental setups we evaluate \methodname on:} \textbf{(a)} the ``put sushi in a metallic pot'' task which requires retargeting, \textbf{(b)} the task of opening an unseen door, and \textbf{(c)} fine-tuning on several novel target tasks in a held out toykitchen environment.}}
  \vspace{-0.6cm}
  \label{fig:experiments}
\end{figure*}

\vspace{0.1cm}
\section{Experimental Evaluation of \methodname and Takeaways for Robotic RL}
\label{sec:result}
\vspace{0.1cm}
The goal of our experiments is to validate if \methodname\ can learn effective policies from only a handful of user-provided demonstrations for a target task, by effectively utilizing previously-collected robotic datasets for pre-training. We also aim to understand whether the design decisions introduced in Section~\ref{sec:design_choices} are crucial for attaining good robotic manipulation performance. To this end, we evaluate \methodname\ in a variety of robotic manipulation settings, and compare it to state of the art methods, which either do not use offline RL or do not learn end-to-end by employing some form of visual representation learning. We evaluate in three scenarios: \textbf{(a)} when the target task requires retargeting the behavior of an existing skill, in this case changing the type of object types it interacts with, \textbf{(b)} when the target task requires performing a previously observed task but this time in a previously unseen domain, and \textbf{(c)} when the target task requires learning a new skill in a new domain, by using the target demonstrations. We also perform a diagnostic study in simulation in Appendix~\ref{app:sim_diagnostic} (Table~\ref{tab:sim_complete}).

\vspace{0.1cm}
\subsection{Setup and Comparisons}
\vspace{0.1cm}

\textbf{Real-world experimental setup.} We directly utilize the publicly available \emph{Bridge Dataset}
\cite{ebert2021bridge} for pre-training, as it provides a large number of robot demonstrations for a diverse set of tasks in multiple domains, i.e., multiple different toy kitchens. We use the same WidowX250 robot platform for our evaluations. The bridge dataset contains distinct tasks, each differing in terms of the objects that the robot interacts with and the domain the task is situated in. We assign a different task identifier to each task in the dataset for pre-training. We also evaluate on an additional door-opening task not present in the Bridge Dataset, where we collected demonstrations for opening and closing a variety of doors, and test our system on new, unseen doors. {More details are in Appendix~\ref{app:exp_setup}.}

\textbf{Comparisons.} Since the datasets we use (both the pre-training bridge dataset from \citep{ebert2021bridge} and the newly collected door opening data) consist of human demonstrations, as indicated by prior work~\citep{mandlekar2021what}, the strongest prior method in this setting is behavioral cloning (BC), which attempts to simply imitate the action of the demonstrator based on the current state. We incorporate BC in a pipeline similar to \methodname, denoted as \textbf{BC (finetune)}, where we first run BC on the pre-training dataset, and then finetune it using the demonstrations on the target task using the same batch mixing as in \methodname. 
To ensure that our BC baselines are well-tuned, we utilize standard practices of cross-validation via a held-out validation set to tune hyperparameters and make early stopping decisions are we elaborate on in {Appendix~\ref{app:hyperparams}}.
Next, to assess the importance of performing pre-training \emph{followed} by fine-tuning, we compare \methodname to \textbf{(i)} jointly training on the pre-training and fine-tuning data with CQL, which is equivalent to the COG approach of \citet{singh2020cog},
\textbf{(ii)} multi-task offline CQL (\textbf{CQL (zero-shot)}) that does not use the target demonstrations at all, and \textbf{(iii)} utilizing CQL to train on target demonstrations alone from scratch, with no pre-training data included (\textbf{CQL (target data only)}).
We also make the analogous comparison for BC, jointly training BC on the pre-training and target task data from scratch (\textbf{BC (joint)}) which is equivalent to \citep{ebert2021bridge}. For fairness of comparison, BC, CQL, and \methodname (both for zero-shot, joint-training and fine-tuning) use the \emph{same} exact architecture, including our learned-spatial embedding described in Section~\ref{sec:design_choices}.

\vspace{0.1cm}
\subsection{Experimental Results}
\vspace{0.1cm}

\begin{table}[h]
\centering
\resizebox{0.5\linewidth}{!}{\begin{tabular}{l|r}
\toprule
\textbf{Method} & \textbf{Success rate}\\ \midrule
BC (zero-shot) & 0/30 \\
BC (finetune) & 0/30  \\ 
CQL (zero-shot) & 2/30 \\
\midrule
\textbf{\methodname (Ours)} & \textbf{14/30} \\ 
\bottomrule
\end{tabular}
}
\vspace{-0.2cm}
\caption{\footnotesize{\textbf{Performance of \methodname for ``put sushi in metallic pot'' in Scenario 1.} \methodname substantially outperforms BC (finetune), even though it is provided access to only demonstration data. We also show some examples comparing some trajectories of BC and \methodname in Appendix ~\ref{app:exp_results}.}}
\label{tab:retarget}
\vspace{-0.2cm}
\end{table}

\begin{table*}[h]
\centering
\resizebox{0.65\linewidth}{!}{
\begin{tabular}{c|c||c|c|c|c|c|c|c}
    \toprule
    & & & \multicolumn{2}{c|}{\textbf{zero-shot}} &  \multicolumn{2}{c|}{\textbf{Joint Training}} & \multicolumn{2}{c}{\textbf{Target data only}} \\
    \midrule
      {\textbf{Task}} & \textbf{\methodname (Ours)} & \textbf{BC (fine.)}   & \textbf{CQL} & \textbf{BC} & \textbf{COG} & \textbf{BC}& \textbf{CQL} & \textbf{BC} \\
    \midrule
   Open Door & \textbf{12/20} & 10/20 & 0/20 & 0/20 & 5/20  & 7/20 &  4/20  & 7/20  \\
    \bottomrule
    \end{tabular}
}
\caption{\footnotesize{{\textbf{Successes vs. total trials for opening a new target door in Scenario 2}}. \methodname outperforms both BC (finetune) and BC (joint) given access to the same data. Note that joint training is worse than finetuning from the pre-trained initialization.}}
\label{tab:adapting}
\vspace{-0.5cm}
\end{table*}

\textbf{Scenario 1: Re-targeting skills for existing tasks to handle new objects.} We utilized the subset of the bridge data with pick-and-place tasks in one toy kitchen for pre-training, and selected the ``put sushi in pot'' task as our target task. This task is depicted in the bridge dataset, but only using an orange transparent pot (see \autoref{fig:experiments} (a)). In order to construct a scenario where the offline policy at the end of pre-training must be re-targeted to act on a different object, we collected only \emph{ten} demonstrations that place the sushi in a metallic pot and used these demonstrations for fine-tuning.
This scenario is challenging since the metallic pot differs significantly from the orange transparent pot visually. 
By pre-training on all pick-and-place tasks in this domain (32 tasks) and fine-tuning on this data and 10 demonstrations, \methodname\ is able to obtain a policy that is re-targeted towards the metal pot. On the other hand, the policy learned by BC confuses arbitrary patches on the tabletop with the pot. Quantitatively, observe in \autoref{tab:retarget} that \methodname is able to complete the task with reasonable accuracy across a set of easy and hard initial positions, whereas zero-shot and fine-tuned BC are completely unable to solve the task. The fact that zero-shot CQL has difficulty solving the task indicates that target demonstrations are necessary, and \methodname is able to attain successful behavior with just ten demonstrations.

\textbf{Scenario 2: Generalizing to previously unseen domains.} Next, we study whether \methodname can adapt behaviors seen in the pre-training data to new domains. We study a door opening task, which requires significantly more complex maneuvers and precise control compared to the pick-and-place tasks from above (as seen in the video present in the supplementary material and our \hyperlink{https://sites.google.com/view/ptr-final/home}{website}).
The doors in the pre-training data exhibit different sizes, shapes, handle types and visual appearances, and the target door (shown in \autoref{fig:experiments}(b)) we wish to open and the corresponding toy kitchen domain are never seen previously in the pre-training data. Concretely, for pre-training, we used a dataset of 800 door-opening demonstrations on 12 different doors in 4 different toy kitchen domains, and we utilize 15 demonstrations on a held-out door for fine-tuning. \autoref{tab:adapting} shows that \methodname improves over both BC baselines and joint training with CQL (or COG). Due to the limited target data and the associated task complexity, in order to succeed, an method must effectively leverage the pre-training data to learn a general policy that attempts to solve the task, and then specialize it to the target door.

Interestingly, \autoref{tab:adapting} shows that while jointly training on the pre-training and fine-tuning data (or COG~\citep{singh2020cog}) by itself does not outperform BC (joint), the pre-training and fine-tuning approach in \methodname{} leads to significantly better performance, improving over the best BC approach. Since CQL (joint) is equivalent to \methodname{}, but with no Phase 1, this large performance gap indicates the efficacy of offline RL methods trained on large diverse datasets at providing good initializations for learning new downstream tasks. We believe that this finding may be of independent interest to robotic offline RL practitioners: when utilizing multi-task offline RL, it might be better first to run multi-task pre-training followed by fine-tuning, as opposed to jointly training from scratch.

\begin{table*}
    \centering
    \resizebox{0.9\linewidth}{!}{\begin{tabular}{c||c||c|cc|cc|cc|c}
    \toprule
    & &  \multicolumn{3}{c|}{\textbf{BC finetuning}} & \multicolumn{2}{c|}{\textbf{Joint training}} &  \multicolumn{2}{c|}{\textbf{Target data only}} &  {\textbf{Meta-learning}} \\
    \midrule
      {\textbf{Task}} & \textbf{\methodname (Ours)} & \textbf{BC (fine.)} & \textbf{Autoreg. BC}& \textbf{BeT} &\textbf{{COG}} & \textbf{{BC}} & \textbf{CQL} & \textbf{BC} & {\textbf{MACAW}} \\
    \midrule
    Take croissant from metal bowl & \textbf{7/10} & 3/10 & 5/10 & 1/10 & 4/10 & 4/10 & 0/10 & 1/10  & 0/10 \\
    Put sweet potato on plate & \textbf{7/20} & 1/20 & 1/20 & 0/20 & 0/20 & 0/20 & 0/20 & 0/20 & 0/20 \\
    Place knife in pot & \textbf{4/10} & 2/10 & 2/10 & 0/10 & 1/10 & 3/10 & 3/10 & 0/10 & 0/10 \\
    Put cucumber in pot & \textbf{5/10} & 0/10 & 1/10 & 0/10 & 2/10 & 1/10 & 0/10 & 0/10 & 0/10 \\
    \bottomrule
    \end{tabular}}
    \caption{\footnotesize{\textbf{Performance of \methodname and other baseline methods for new tasks in Scenario 3.} Note that \methodname outperforms all other baselines including BC (finetune), BC with more expressive policy classes (BeT~\citep{shafiullah2022behavior}, Auto-regressive), offline RL with no pre-training (``Target data only'') and joint training~\citep{singh2020cog,ebert2021bridge}. PTR also outperforms few-shot gradient-based meta learning methods such as MACAW~\citep{mitchell2021offline}, which fail to attain non-zero performance.}}
    \label{tab:scenario4}
\end{table*}

\textbf{Scenario 3: Learning to solve new tasks in new domains.} 
Finally, we evaluate the efficacy of \methodname in learning to solve a new task in a new domain. 
This scenario presents a generalization requirement that is significantly more challenging than the previously studied scenarios, since both the task and the domain are never seen before. This task is represented via a new task identifier, and pre-training receives no data for this task identifier, or even any data from the kitchen where this task is situated. We pre-train on all 80 pick-and-place style tasks from the bridge dataset, while holding out any data from the new task kitchen, and then fine-tune on 10 demonstrations for 4 target tasks independently in this new kitchen, as shown in Table~\ref{tab:scenario4}. 
Methods that utilize more expressive policy architectures (an auto-regressive policy or behavior transformers (BeT)~\citep{shafiullah2022behavior}) do not lead to improved performance compared to the standard BC (finetune) approach, and we find that PTR outperforms these approaches. Please find more details on the implementation of auto-regressive BC and BeT in Section \ref{app:hyperparams}. This might appear surprising, and perhaps just a hyper-parameter tuning artifact at first, but we present additional qualitative and quantitative analysis aiming at understanding the reasons behind why our offline RL-based PTR approach works better in Section~\ref{sec:rl_vs_bc}. We also compare to MACAW~\citep{mitchell2021offline}, an offline meta-RL method that utilizes advantage-weighted regression~\citep{peng2019awr} for gradient-based few-shot adaptation, and find that this approach is unable to learn policies that succeed. We discuss the hyperparameter configurations that we tried for this approach in Appendix \ref{app:scenario4}. Finally, observe in Table~\ref{tab:scenario4} that joint training with CQL or BC, or just using target data, without any pre-training for CQL or BC, all perform significantly worse than PTR.

\begin{table}
    \centering
    \resizebox{0.8\linewidth}{!}{\begin{tabular}{c||c||cc}
    \toprule
    & &  \multicolumn{2}{c}{\textbf{Pre-train. rep. + BC finetune}} \\
    \midrule
      {\textbf{Task}} & \textbf{\methodname (Ours)} & \textbf{R3M} & \textbf{MAE} \\
    \midrule
    Take croissant from bowl & \textbf{7/10} & 1/10 & 3/10 \\
    Put sweet potato on plate & \textbf{7/20} & 0/20 & 1/20  \\
    Place knife in pot & \textbf{4/10}  & 0/10 & 0/10  \\
    Put cucumber in pot & \textbf{5/10}  & 0/10 & 0/10 \\
    \bottomrule
    \end{tabular}}
    \caption{\footnotesize{\textbf{Performance of \methodname and other pre-training methods (R3M and MAE).} While both R3M~\citep{nair2022r3m} and MAE~\citep{xiao2022masked} help improve performance over na\"ively applying BC on the target data, \methodname outperforms both.}}
    \label{tab:rep_learning_comparison}
    \vspace{-0.6cm}
\end{table}

\vspace{0.05cm}
\subsection{Comparison to non-RL Visual Pre-Training Methods}
\vspace{0.05cm}

We also compare PTR to approaches that utilize the diverse bridge dataset or Internet-scale data
for task-agnostic visual representation learning, followed by down-stream behavioral cloning only on the target fine-tuning task which utilizes the representation learned during pre-training. In particular, we compare to two approaches: R3M~\citep{nair2022r3m}, which utilizes the Ego4D dataset of human videos to obtain a representation, and MVP~\citep{radosavovic2022real,xiao2022masked}, which trains a masked auto-encoder~\citep{he2111masked} on the Bridge Dataset and utilizes the learned latent space as the representation of the new image. Observe in Table~\ref{tab:rep_learning_comparison} that, while utilizing R3M or MAE does improve over running BC on the target data alone (compare R3M and MAE in Table~\ref{tab:rep_learning_comparison} to BC on target data only in Table~\ref{tab:scenario4}), the pre-training scheme from PTR outperforms both of these prior pre-training approaches, indicating the efficacy of offline RL pre-training on diverse robot data in recovering useful representations for downstream policy learning.

\vspace{0.1cm}
\subsection{Understanding the Benefits of PTR over BC}
\label{sec:rl_vs_bc}
\vspace{0.1cm}

One natural question to ask given the results in this paper is: why does utilizing an offline RL method for pre-training and fine-tuning as in \methodname outperform BC-based methods even though the dataset is quite ``BC-friendly'', consisting of only demonstrations? The answer to this question is not obvious, especially since joint training with BC still outperforms jointly training with CQL on both pre-training and target demonstration data (COG) in our results in Table~\ref{tab:scenario4}.  

\begin{figure}[h]
\centering
\vspace{-0.4cm}
  \includegraphics[width=0.8\linewidth]{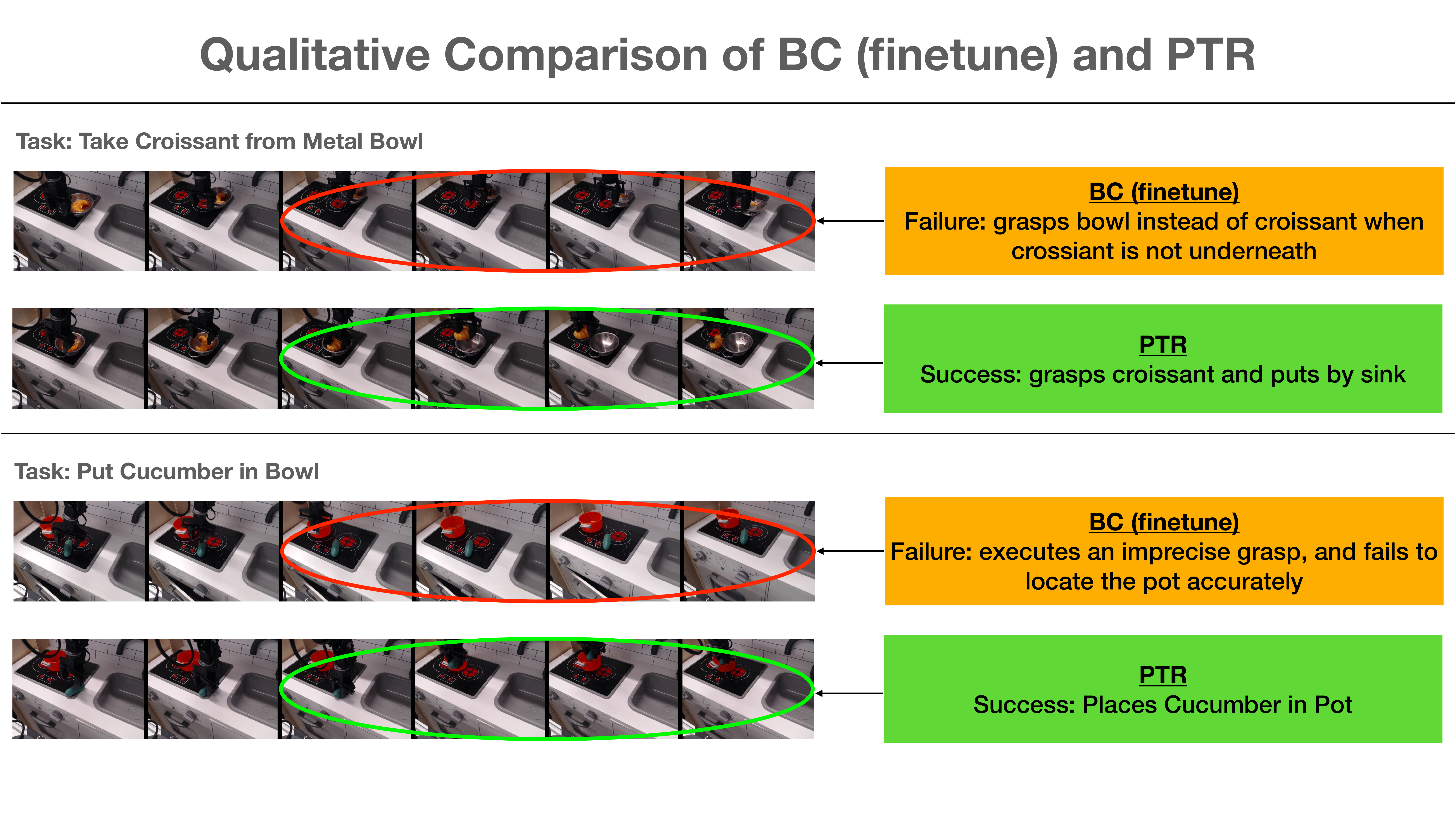}
  \vspace{-0.3cm}
  \caption{\footnotesize \textbf{Qualitative successes of \methodname visualized alongside failures of BC (fine-tune).} As an example, observe that while \methodname is accurately able to reach to the croissant and grasp it to solve the task, BC (finetune) is imprecise and grasps the bowl instead of the croissant resulting in failure.}
  \label{fig:dumb_behavior}
  \vspace{-0.3cm}
\end{figure}

To understand the reason behind improvements from RL, we perform a qualitative evaluation of the policies learned by PTR and BC (finetune) on two tasks: take croissant from metal bowl and put cucumber in bowl in Figure~\ref{fig:dumb_behavior}. We find that the failure mode of BC policies can be primarily explained as a lack of precision in locating the object, or a prematurely-executed grasping action. This is especially prevalent in settings where the object of interest is farther away from the robot gripper at the initial state, and hints at the inability of BC to prioritize learning the critical decisions (e.g., precisely moving over the object before the grasping action) over non-critical ones (e.g., the action to take to reach nearby the object from farther away). On the other hand, RL can learn to make such critical decisions correctly as shown in Figure~\ref{fig:dumb_behavior}. We present additional rollouts in Appendix~\ref{app:exp_setup}.

\begin{table}[h]
\centering
\resizebox{0.75\linewidth}{!}{\begin{tabular}{c|r|r||r}
\toprule
\textbf{Task} & \textbf{BC (finetune)} & \textbf{\methodname} &\textbf{AW-BC (finetune)}  \\ \midrule
Cucumber &  0/10 & 5/10 &  {5/10} \\
Croissant & 3/10 &  7/10  & {6/10} \\
\bottomrule
\end{tabular}}
\caption{\footnotesize{\textbf{Performance of advantage-weighted BC} on two tasks from Table~\ref{tab:scenario4}. Observe that weighting the BC objective using advantage estimates from the Q-function learned by \methodname leads to much better performance than standard BC (finetune), almost recovering PTR performance. This test indicates that the Q-function in \methodname allows us to be accurate on the more critical decisions, thereby preventing the failures of BC.}}
\vspace{-0.3cm}
\label{tab:aw_bc}
\end{table}

Next, to verify if the performance benefits can be explained by the ability of Q-learning to prioritize critical decisions, we run a form of weighted behavioral cloning, where the weights $w_\phi(\bs, \ba)$ are derived from the \emph{advantage estimates computed using a frozen Q-function learned by PTR} after fine-tuning:
\begin{align*}
    w_\phi(\bs, \ba)  \propto \exp(Q_\phi(\bs, \ba) - \max_{\ba'} Q_\phi(\bs, \ba')).
\end{align*}
Note that this is not the same as standard advantage-weighted regression~\citep{peng2019awr}, which uses Monte-Carlo return estimates for computing advantage weights instead of using advantages computed under a Q-function trained via PTR or CQL. As shown in Table~\ref{tab:aw_bc}, we find that this advantage-weighted BC (AW-BC) approach performs significantly better than BC (finetune) method and comparably to PTR, for two tasks (croissant and cucumber from Table~\ref{tab:scenario4}. Since AW-BC is essentially the same as BC, just with a modified weight to indicate the importance of any transition, this performance improvement clearly indicates the benefits of learning value functions via PTR in a pre-training then fine-tuning setting, even when we only have demonstration data.  Note that since AW-BC uses the PTR-derived weights after fine-tuning, it cannot serve as an independent method, but rather amounts to another way to use the PTR value function.

\subsection{Effective Use of High-Capacity Neural Networks}
\vspace{0.1cm}

\begin{figure}
\centering
\vspace{-0.7cm}
\includegraphics[width=0.6\linewidth]{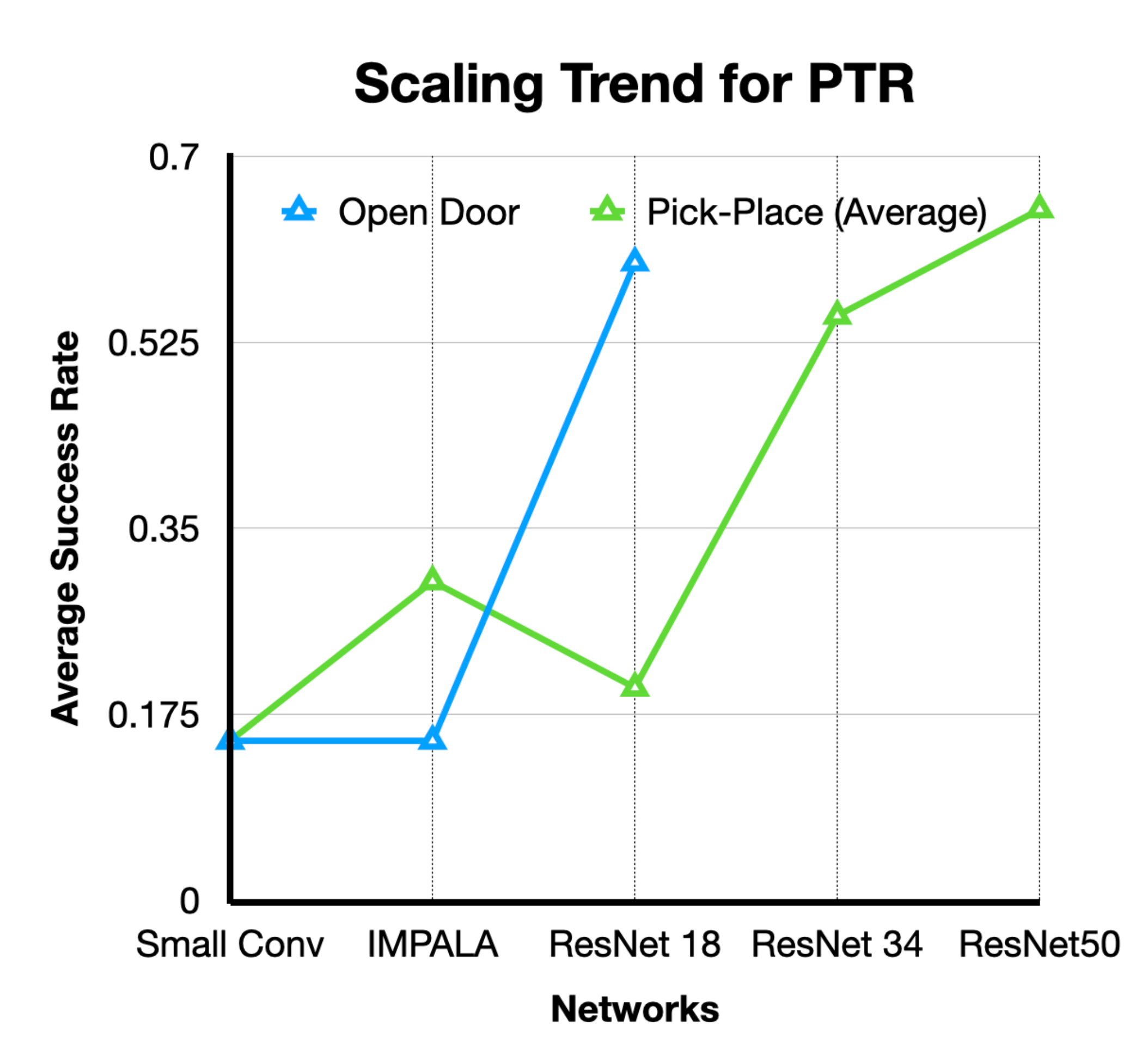}
\vspace{-0.24cm}
\caption{\footnotesize{\label{fig:scaling_ptr} \textbf{Scaling trends for \methodname} on the open door task, and average over two pick and place tasks from Scenario 3. Note that with our design decisions, PTR is able to effectively benefit from high capacity networks.}}
\vspace{-0.6cm}
\end{figure}
To understand the importance of designing techniques that enable us to use high-capacity models for offline RL, we examine the efficacy of PTR with different neural network architectures on the open door task from Scenario 2, and the put cucumber in pot and take croissant out of metallic bowl tasks from Scenario 3. We compare to standard three-layer convolutional network architectures used by prior work for Deepmind control suite tasks (see for example, \citet{kostrikov2020image}), an IMPALA~\citep{espeholt2018impala} ResNet that consists of 15 convolutional layers spread across a stack of 3 residual blocks, and the ResNet 18, 34, and 50 architectures with our proposed design decisions. Observe in Figure~\ref{fig:scaling_ptr} that the performance of smaller networks (Small, IMPALA) is significantly worse than the ResNet in the door opening task. For the pick-and-place tasks that contain a much larger dataset, Small, IMPALA and ResNet18 all perform much worse than ResNet 34 and ResNet 50. In Appendix~\ref{app:design} we show that ResNet 34 models perform much worse if our prescribed design decisions are not used.

\vspace{0.1cm}
\subsection{Autonomous Online Fine-Tuning}
\label{sec:experiments_online}
\vspace{0.1cm}

\begin{table}
\centering
\resizebox{0.8\linewidth}{!}{\begin{tabular}{c|c|c}
\toprule
& \textbf{SACfD} &  \textbf{\methodname (offline $\rightarrow$ online)}  \\ \midrule
All positions & 0\% $\rightarrow$ 0\%   &  \textbf{53\%  $\rightarrow$ 73\%} \\
Novel OOD positions & 0\% $\rightarrow$ 0\%  & \textbf{13\%  $\rightarrow$  60\%} \\
\bottomrule
\end{tabular}}
\caption{\footnotesize{\textbf{Performance before and after online fine-tuning.} The success rate of the \methodname pre-trained policy is improved significantly from online fine-tuning, especially on novel out-of-distribution (OOD) initial positions that must be learned entirely from autonomous interaction in the real world. The results are reported as the average of 3 trials from each initial position.}}
\vspace{-0.3cm}
\label{tab:online-finetune}
\end{table}

So far, we've evaluated \methodname with offline fine-tuning to new tasks. However, by pre-training representations with offline RL, we can also enable autonomous improvement through online RL fine-tuning. In this section, we will demonstrate this benefit by showing that an offline initialization learned by PTR pre-training can be effectively fine-tuned autonomously with online rollouts. This procedure provides a way forward to build self-improving robotic RL systems that bring the best of diverse robotic datasets and learning via online interaction.

\textbf{Task.} For this experiment, we consider the ``open door'' task from Scenario 2. Our goal is to improve the success rate of the learned policy obtained after PTR pre-training and offline fine-tuning using autonomous online rollouts from ten initial positions. These ten initial positions consist of five positions obtained by randomly sampling from the target demonstrations used for offline fine-tuning, and five more challenging \textbf{out-of-distribution initial positions}, that were never seen before.

\begin{figure}[t]
\centering
\includegraphics[width=0.6\linewidth]{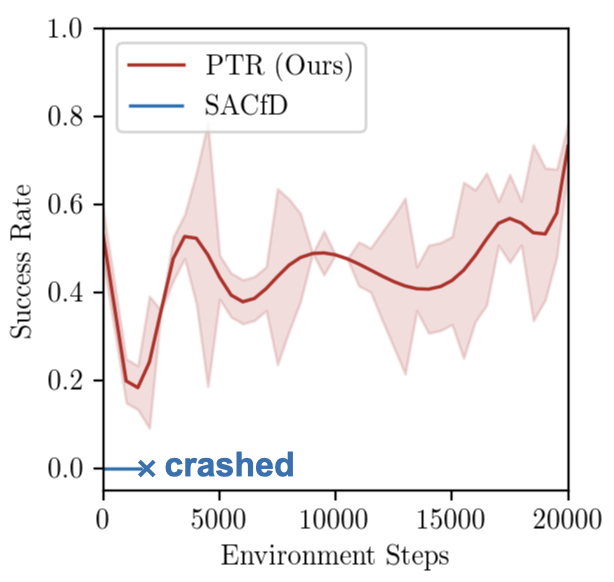}
\vspace{-0.24cm}
\caption{\footnotesize{\label{fig:online_door} \textbf{Online fine-tuning for \methodname} on the open door task. \methodname improves the success rate of the pre-trained policy from 53\% to 73\% (from 13\% to 60\% for the harder positions), while SACfD crashes due to unsafe behavior during exploration. We ran \methodname online fine-tuning for 2 seeds in the real world.}}
\vspace{-0.6cm}
\end{figure}

\textbf{Reward functions.} To run RL, we need a mechanism to annotate every online rollout with a reward signal. Following prior works~\citep{singh2019, kalashnikov2021mt}, we trained a neural-network binary classifier to detect a given visual observation as a success (+1 reward) or failure (0 reward) and use it to annotate rollouts executed during online interaction. 

\textbf{Reset policy.} To run online fine-tuning autonomously without any human intervention in the real world, we also need a ``reset policy'' that closes the door after a successful online rollout. To this end, we also pre-trained a close-door policy separately, which is used only for resetting the door. Note that online fine-tuning only fine-tunes the open-door policy, while the reset policy is kept fixed throughout.

\textbf{Online training setup.} Equipped with the reset policy and the reward classifier, we are able to run online fine-tuning in the real world. Starting from the pre-trained policy obtained via PTR, our method alternates between collecting a new trajectory and taking gradient steps. The update-to-data ratio~\citep{2021arXiv210105982C} is set to 10, which means that we  make 10 gradient updates for every environment step. More details about our implementation and evaluations can be found in Appendix~\ref{app:online_finetuning}.

\begin{figure}
\vspace{-0.4cm}
\centering
\includegraphics[width=0.85\linewidth]{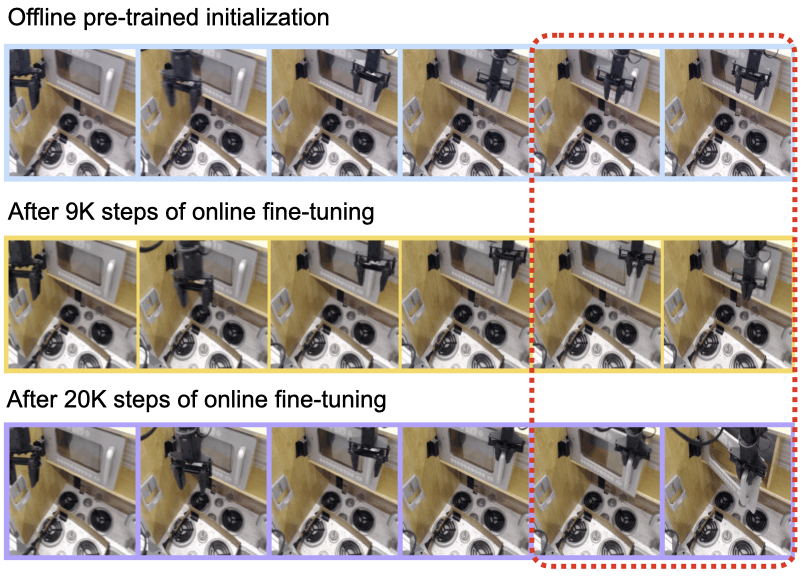}
\vspace{-0.24cm}
\caption{\footnotesize{\label{fig:online-improvement} \textbf{Evolution of learned behaviors during autonomous online fine-tuning of PTR starting from one of the hard initial positions.} The blue box illustrates that the offline initialization fails to grasp the handle. After 9K steps of online interaction, it successfully grasps the handle but fails to open the door. After 20K steps, it learns to successfully open the door.}}
\vspace{-0.4cm}
\end{figure}

\textbf{Results.} We compare our method with a prior method that trains SAC~\citep{haarnoja2018soft} from scratch using both online data and offline demonstrations (denoted by ``SACfD''). This approach is an improved version of DDPGfD~\citep{vecerik2017leveraging} which uses a stronger off-policy RL algorithm (SAC). We present the learning curve during the online fine-tuning in Figure~\ref{fig:online_door}, and the success rates before and after fine-tuning in Table~\ref{tab:online-finetune}. As shown in Figure~\ref{fig:online_door}, it was difficult to run SACfD over a long time on the robot, as the system crashes due to unsafe actions during exploration (pictures shown in Appendix~\ref{app:online_finetuning}). In contrast, the pre-trained PTR policy is able to perform online exploration in a stable manner, and improve the success rate of the pre-trained policy within 20K steps of online interaction. Specifically, this boost in performance stems from learning to solve the task from \textbf{3/5} of the more challenging, out-of-distribution initial positions,
that were never seen before in the prior data, as shown in Figure~\ref{fig:online-improvement}. Overall, our results show the efficacy of PTR as a general-purpose pre-training paradigm for robotic RL.

%% file: main.bbl
\begin{thebibliography}{62}
\providecommand{\natexlab}[1]{#1}
\providecommand{\url}[1]{\texttt{#1}}
\expandafter\ifx\csname urlstyle\endcsname\relax
  \providecommand{\doi}[1]{doi: #1}\else
  \providecommand{\doi}{doi: \begingroup \urlstyle{rm}\Url}\fi

\bibitem[Ahn et~al.(2022)Ahn, Brohan, Brown, Chebotar, Cortes, David, Finn,
  Gopalakrishnan, Hausman, Herzog, et~al.]{ahn2022can}
Michael Ahn, Anthony Brohan, Noah Brown, Yevgen Chebotar, Omar Cortes, Byron
  David, Chelsea Finn, Keerthana Gopalakrishnan, Karol Hausman, Alex Herzog,
  et~al.
\newblock Do as i can, not as i say: Grounding language in robotic affordances.
\newblock \emph{arXiv preprint arXiv:2204.01691}, 2022.

\bibitem[Andrychowicz et~al.(2017)Andrychowicz, Wolski, Ray, Schneider, Fong,
  Welinder, McGrew, Tobin, Pieter~Abbeel, and
  Zaremba]{andrychowicz2017hindsight}
Marcin Andrychowicz, Filip Wolski, Alex Ray, Jonas Schneider, Rachel Fong,
  Peter Welinder, Bob McGrew, Josh Tobin, OpenAI Pieter~Abbeel, and Wojciech
  Zaremba.
\newblock Hindsight experience replay.
\newblock \emph{Advances in neural information processing systems}, 30, 2017.

\bibitem[{Bhatt} et~al.(2019){Bhatt}, {Argus}, {Amiranashvili}, and
  {Brox}]{2019arXiv190205605B}
Aditya {Bhatt}, Max {Argus}, Artemij {Amiranashvili}, and Thomas {Brox}.
\newblock {CrossNorm: Normalization for Off-Policy TD Reinforcement Learning}.
\newblock \emph{arXiv e-prints}, art. arXiv:1902.05605, February 2019.

\bibitem[Bjorck et~al.(2021)Bjorck, Gomes, and Weinberger]{bjorck2021towards}
Johan Bjorck, Carla~P Gomes, and Kilian~Q Weinberger.
\newblock Towards deeper deep reinforcement learning.
\newblock \emph{arXiv preprint arXiv:2106.01151}, 2021.

\bibitem[Brown et~al.(2020)Brown, Mann, Ryder, Subbiah, Kaplan, Dhariwal,
  Neelakantan, Shyam, Sastry, Askell, et~al.]{brown2020language}
Tom Brown, Benjamin Mann, Nick Ryder, Melanie Subbiah, Jared~D Kaplan, Prafulla
  Dhariwal, Arvind Neelakantan, Pranav Shyam, Girish Sastry, Amanda Askell,
  et~al.
\newblock Language models are few-shot learners.
\newblock \emph{Advances in neural information processing systems},
  33:\penalty0 1877--1901, 2020.

\bibitem[Chebotar et~al.(2021)Chebotar, Hausman, Lu, Xiao, Kalashnikov, Varley,
  Irpan, Eysenbach, Julian, Finn, et~al.]{chebotar2021actionable}
Yevgen Chebotar, Karol Hausman, Yao Lu, Ted Xiao, Dmitry Kalashnikov, Jake
  Varley, Alex Irpan, Benjamin Eysenbach, Ryan Julian, Chelsea Finn, et~al.
\newblock Actionable models: Unsupervised offline reinforcement learning of
  robotic skills.
\newblock \emph{arXiv preprint arXiv:2104.07749}, 2021.

\bibitem[{Chen} et~al.(2021){Chen}, {Wang}, {Zhou}, and
  {Ross}]{2021arXiv210105982C}
Xinyue {Chen}, Che {Wang}, Zijian {Zhou}, and Keith {Ross}.
\newblock {Randomized Ensembled Double Q-Learning: Learning Fast Without a
  Model}.
\newblock \emph{arXiv e-prints}, art. arXiv:2101.05982, January 2021.
\newblock \doi{10.48550/arXiv.2101.05982}.

\bibitem[Dorfman and Tamar(2020)]{dorfman2020offline}
Ron Dorfman and Aviv Tamar.
\newblock Offline meta reinforcement learning.
\newblock \emph{arXiv e-prints}, pages arXiv--2008, 2020.

\bibitem[Ebert et~al.(2021)Ebert, Yang, Schmeckpeper, Bucher, Georgakis,
  Daniilidis, Finn, and Levine]{ebert2021bridge}
Frederik Ebert, Yanlai Yang, Karl Schmeckpeper, Bernadette Bucher, Georgios
  Georgakis, Kostas Daniilidis, Chelsea Finn, and Sergey Levine.
\newblock Bridge data: Boosting generalization of robotic skills with
  cross-domain datasets.
\newblock \emph{arXiv preprint arXiv:2109.13396}, 2021.

\bibitem[Emmons et~al.(2021)Emmons, Eysenbach, Kostrikov, and
  Levine]{emmons2021rvs}
Scott Emmons, Benjamin Eysenbach, Ilya Kostrikov, and Sergey Levine.
\newblock Rvs: What is essential for offline rl via supervised learning?
\newblock \emph{arXiv preprint arXiv:2112.10751}, 2021.

\bibitem[Espeholt et~al.(2018)Espeholt, Soyer, Munos, Simonyan, Mnih, Ward,
  Doron, Firoiu, Harley, Dunning, et~al.]{espeholt2018impala}
Lasse Espeholt, Hubert Soyer, Remi Munos, Karen Simonyan, Vlad Mnih, Tom Ward,
  Yotam Doron, Vlad Firoiu, Tim Harley, Iain Dunning, et~al.
\newblock Impala: Scalable distributed deep-rl with importance weighted
  actor-learner architectures.
\newblock In \emph{International Conference on Machine Learning}, pages
  1407--1416. PMLR, 2018.

\bibitem[Eysenbach et~al.(2021)Eysenbach, Levine, and
  Salakhutdinov]{eysenbach2021replacing}
Ben Eysenbach, Sergey Levine, and Russ~R Salakhutdinov.
\newblock Replacing rewards with examples: Example-based policy search via
  recursive classification.
\newblock \emph{Advances in Neural Information Processing Systems},
  34:\penalty0 11541--11552, 2021.

\bibitem[Fujimoto and Gu(2021)]{fujimoto2021minimalist}
Scott Fujimoto and Shixiang~Shane Gu.
\newblock A minimalist approach to offline reinforcement learning.
\newblock \emph{arXiv preprint arXiv:2106.06860}, 2021.

\bibitem[Fujimoto et~al.(2018)Fujimoto, Meger, and Precup]{fujimoto2018off}
Scott Fujimoto, David Meger, and Doina Precup.
\newblock Off-policy deep reinforcement learning without exploration.
\newblock \emph{arXiv preprint arXiv:1812.02900}, 2018.

\bibitem[Haarnoja et~al.(2018)Haarnoja, Zhou, Abbeel, and
  Levine]{haarnoja2018soft}
Tuomas Haarnoja, Aurick Zhou, Pieter Abbeel, and Sergey Levine.
\newblock Soft actor-critic: Off-policy maximum entropy deep reinforcement
  learning with a stochastic actor.
\newblock In \emph{International conference on machine learning}, pages
  1861--1870. PMLR, 2018.

\bibitem[He et~al.(2021)He, Chen, Xie, Li, Doll{\'a}r, and
  Girshick]{he2111masked}
K~He, X~Chen, S~Xie, Y~Li, P~Doll{\'a}r, and RB~Girshick.
\newblock Masked autoencoders are scalable vision learners. arxiv. 2021 doi:
  10.48550.
\newblock \emph{arXiv preprint arXiv.2111.06377}, 2021.

\bibitem[He et~al.(2016)He, Zhang, Ren, and Sun]{resnet}
Kaiming He, Xiangyu Zhang, Shaoqing Ren, and Jian Sun.
\newblock Deep residual learning for image recognition.
\newblock In \emph{Proceedings of the IEEE conference on computer vision and
  pattern recognition}, pages 770--778, 2016.

\bibitem[Hessel et~al.(2019)Hessel, Soyer, Espeholt, Czarnecki, Schmitt, and
  van Hasselt]{hessel2019multi}
Matteo Hessel, Hubert Soyer, Lasse Espeholt, Wojciech Czarnecki, Simon Schmitt,
  and Hado van Hasselt.
\newblock Multi-task deep reinforcement learning with popart.
\newblock In \emph{Proceedings of the AAAI Conference on Artificial
  Intelligence}, volume~33, pages 3796--3803, 2019.

\bibitem[Jaques et~al.(2019)Jaques, Ghandeharioun, Shen, Ferguson, Lapedriza,
  Jones, Gu, and Picard]{jaques2019way}
Natasha Jaques, Asma Ghandeharioun, Judy~Hanwen Shen, Craig Ferguson, Agata
  Lapedriza, Noah Jones, Shixiang Gu, and Rosalind Picard.
\newblock Way off-policy batch deep reinforcement learning of implicit human
  preferences in dialog.
\newblock \emph{arXiv preprint arXiv:1907.00456}, 2019.

\bibitem[Julian et~al.(2020)Julian, Swanson, Sukhatme, Levine, Finn, and
  Hausman]{julian2020never}
Ryan Julian, Benjamin Swanson, Gaurav~S Sukhatme, Sergey Levine, Chelsea Finn,
  and Karol Hausman.
\newblock Never stop learning: The effectiveness of fine-tuning in robotic
  reinforcement learning.
\newblock \emph{arXiv preprint arXiv:2004.10190}, 2020.

\bibitem[Kalashnikov et~al.(2018)Kalashnikov, Irpan, Pastor, Ibarz, Herzog,
  Jang, Quillen, Holly, Kalakrishnan, Vanhoucke, et~al.]{kalashnikov2018qtopt}
Dmitry Kalashnikov, Alex Irpan, Peter Pastor, Julian Ibarz, Alexander Herzog,
  Eric Jang, Deirdre Quillen, Ethan Holly, Mrinal Kalakrishnan, Vincent
  Vanhoucke, et~al.
\newblock Scalable deep reinforcement learning for vision-based robotic
  manipulation.
\newblock In \emph{Conference on Robot Learning}, pages 651--673, 2018.

\bibitem[Kalashnikov et~al.(2021)Kalashnikov, Varley, Chebotar, Swanson,
  Jonschkowski, Finn, Levine, and Hausman]{kalashnikov2021mt}
Dmitry Kalashnikov, Jacob Varley, Yevgen Chebotar, Benjamin Swanson, Rico
  Jonschkowski, Chelsea Finn, Sergey Levine, and Karol Hausman.
\newblock Mt-opt: Continuous multi-task robotic reinforcement learning at
  scale.
\newblock \emph{arXiv preprint arXiv:2104.08212}, 2021.

\bibitem[Kostrikov et~al.(2020)Kostrikov, Yarats, and
  Fergus]{kostrikov2020image}
Ilya Kostrikov, Denis Yarats, and Rob Fergus.
\newblock Image augmentation is all you need: Regularizing deep reinforcement
  learning from pixels.
\newblock \emph{arXiv preprint arXiv:2004.13649}, 2020.

\bibitem[Kostrikov et~al.(2021{\natexlab{a}})Kostrikov, Fergus, Tompson, and
  Nachum]{kostrikov2021offline}
Ilya Kostrikov, Rob Fergus, Jonathan Tompson, and Ofir Nachum.
\newblock Offline reinforcement learning with fisher divergence critic
  regularization.
\newblock In \emph{International Conference on Machine Learning}, pages
  5774--5783. PMLR, 2021{\natexlab{a}}.

\bibitem[Kostrikov et~al.(2021{\natexlab{b}})Kostrikov, Nair, and
  Levine]{kostrikov2021iql}
Ilya Kostrikov, Ashvin Nair, and Sergey Levine.
\newblock Offline reinforcement learning with implicit q-learning.
\newblock 2021{\natexlab{b}}.

\bibitem[Kumar et~al.(2019)Kumar, Fu, Soh, Tucker, and
  Levine]{kumar2019stabilizing}
Aviral Kumar, Justin Fu, Matthew Soh, George Tucker, and Sergey Levine.
\newblock Stabilizing off-policy q-learning via bootstrapping error reduction.
\newblock In \emph{Advances in Neural Information Processing Systems}, pages
  11761--11771, 2019.

\bibitem[Kumar et~al.(2020)Kumar, Zhou, Tucker, and
  Levine]{kumar2020conservative}
Aviral Kumar, Aurick Zhou, George Tucker, and Sergey Levine.
\newblock Conservative q-learning for offline reinforcement learning.
\newblock \emph{arXiv preprint arXiv:2006.04779}, 2020.

\bibitem[Kumar et~al.(2022)Kumar, Hong, Singh, and Levine]{kumar2022should}
Aviral Kumar, Joey Hong, Anikait Singh, and Sergey Levine.
\newblock Should i run offline reinforcement learning or behavioral cloning?
\newblock In \emph{International Conference on Learning Representations}, 2022.
\newblock URL \url{https://openreview.net/forum?id=AP1MKT37rJ}.

\bibitem[Lee et~al.(2022{\natexlab{a}})Lee, Devin, Springenberg, Zhou, Lampe,
  Abdolmaleki, and Bousmalis]{lee2022spend}
Alex~X Lee, Coline Devin, Jost~Tobias Springenberg, Yuxiang Zhou, Thomas Lampe,
  Abbas Abdolmaleki, and Konstantinos Bousmalis.
\newblock How to spend your robot time: Bridging kickstarting and offline
  reinforcement learning for vision-based robotic manipulation.
\newblock \emph{arXiv preprint arXiv:2205.03353}, 2022{\natexlab{a}}.

\bibitem[Lee et~al.(2022{\natexlab{b}})Lee, Nachum, Yang, Lee, Freeman, Xu,
  Guadarrama, Fischer, Jang, Michalewski, et~al.]{lee2022multi}
Kuang-Huei Lee, Ofir Nachum, Mengjiao Yang, Lisa Lee, Daniel Freeman, Winnie
  Xu, Sergio Guadarrama, Ian Fischer, Eric Jang, Henryk Michalewski, et~al.
\newblock Multi-game decision transformers.
\newblock \emph{arXiv preprint arXiv:2205.15241}, 2022{\natexlab{b}}.

\bibitem[Lee et~al.(2022{\natexlab{c}})Lee, Seo, Lee, Abbeel, and
  Shin]{lee2022offline}
Seunghyun Lee, Younggyo Seo, Kimin Lee, Pieter Abbeel, and Jinwoo Shin.
\newblock Offline-to-online reinforcement learning via balanced replay and
  pessimistic q-ensemble.
\newblock In \emph{Conference on Robot Learning}, pages 1702--1712. PMLR,
  2022{\natexlab{c}}.

\bibitem[Levine et~al.(2016)Levine, Finn, Darrell, and Abbeel]{levine2016end}
Sergey Levine, Chelsea Finn, Trevor Darrell, and Pieter Abbeel.
\newblock End-to-end training of deep visuomotor policies.
\newblock \emph{The Journal of Machine Learning Research}, 17\penalty0
  (1):\penalty0 1334--1373, 2016.

\bibitem[Li et~al.(2019)Li, Vuong, Liu, Liu, Ciosek, Ross, Christensen, and
  Su]{li2019multi}
Jiachen Li, Quan Vuong, Shuang Liu, Minghua Liu, Kamil Ciosek, Keith Ross,
  Henrik~Iskov Christensen, and Hao Su.
\newblock Multi-task batch reinforcement learning with metric learning.
\newblock \emph{arXiv preprint arXiv:1909.11373}, 2019.

\bibitem[Lin et~al.(2022)Lin, Wan, Xu, Liang, and Zhang]{lin2022model}
Sen Lin, Jialin Wan, Tengyu Xu, Yingbin Liang, and Junshan Zhang.
\newblock Model-based offline meta-reinforcement learning with regularization.
\newblock \emph{arXiv preprint arXiv:2202.02929}, 2022.

\bibitem[Ma et~al.(2022)Ma, Sodhani, Jayaraman, Bastani, Kumar, and
  Zhang]{ma2022vip}
Yecheng~Jason Ma, Shagun Sodhani, Dinesh Jayaraman, Osbert Bastani, Vikash
  Kumar, and Amy Zhang.
\newblock Vip: Towards universal visual reward and representation via
  value-implicit pre-training.
\newblock \emph{arXiv preprint arXiv:2210.00030}, 2022.

\bibitem[Mandlekar et~al.(2020)Mandlekar, Ramos, Boots, Savarese, Fei-Fei,
  Garg, and Fox]{mandlekar2020iris}
Ajay Mandlekar, Fabio Ramos, Byron Boots, Silvio Savarese, Li~Fei-Fei, Animesh
  Garg, and Dieter Fox.
\newblock Iris: Implicit reinforcement without interaction at scale for
  learning control from offline robot manipulation data.
\newblock In \emph{2020 IEEE International Conference on Robotics and
  Automation (ICRA)}, pages 4414--4420. IEEE, 2020.

\bibitem[Mandlekar et~al.(2021)Mandlekar, Xu, Wong, Nasiriany, Wang, Kulkarni,
  Li, Savarese, Zhu, and Mart{\'\i}n-Mart{\'\i}n]{mandlekar2021what}
Ajay Mandlekar, Danfei Xu, Josiah Wong, Soroush Nasiriany, Chen Wang, Rohun
  Kulkarni, Fei-Fei Li, Silvio Savarese, Yuke Zhu, and Roberto
  Mart{\'\i}n-Mart{\'\i}n.
\newblock What matters in learning from offline human demonstrations for robot
  manipulation.
\newblock In \emph{5th Annual Conference on Robot Learning}, 2021.
\newblock URL \url{https://openreview.net/forum?id=JrsfBJtDFdI}.

\bibitem[{Mitchell} et~al.(2020){Mitchell}, {Rafailov}, {Peng}, {Levine}, and
  {Finn}]{2020arXiv200806043M}
Eric {Mitchell}, Rafael {Rafailov}, Xue~Bin {Peng}, Sergey {Levine}, and
  Chelsea {Finn}.
\newblock {Offline Meta-Reinforcement Learning with Advantage Weighting}.
\newblock \emph{arXiv e-prints}, art. arXiv:2008.06043, August 2020.

\bibitem[Mitchell et~al.(2021)Mitchell, Rafailov, Peng, Levine, and
  Finn]{mitchell2021offline}
Eric Mitchell, Rafael Rafailov, Xue~Bin Peng, Sergey Levine, and Chelsea Finn.
\newblock Offline meta-reinforcement learning with advantage weighting.
\newblock In \emph{International Conference on Machine Learning}, pages
  7780--7791. PMLR, 2021.

\bibitem[Nair et~al.(2020)Nair, Dalal, Gupta, and Levine]{nair2020accelerating}
Ashvin Nair, Murtaza Dalal, Abhishek Gupta, and Sergey Levine.
\newblock Accelerating online reinforcement learning with offline datasets.
\newblock \emph{arXiv preprint arXiv:2006.09359}, 2020.

\bibitem[Nair et~al.(2022)Nair, Rajeswaran, Kumar, Finn, and
  Gupta]{nair2022r3m}
Suraj Nair, Aravind Rajeswaran, Vikash Kumar, Chelsea Finn, and Abhinav Gupta.
\newblock R3m: A universal visual representation for robot manipulation.
\newblock \emph{arXiv preprint arXiv:2203.12601}, 2022.

\bibitem[Nakamoto et~al.(2023)Nakamoto, Zhai, Singh, Mark, Ma, Finn, Kumar, and
  Levine]{nakamoto2023calql}
Mitsuhiko Nakamoto, Yuexiang Zhai, Anikait Singh, Max~Sobol Mark, Yi~Ma,
  Chelsea Finn, Aviral Kumar, and Sergey Levine.
\newblock Cal-{QL}: Calibrated offline rl pre-training for efficient online
  fine-tuning.
\newblock \emph{arXiv preprint arXiv:2303.05479}, 2023.

\bibitem[Parisotto et~al.(2015)Parisotto, Ba, and
  Salakhutdinov]{parisotto2015actor}
Emilio Parisotto, Jimmy~Lei Ba, and Ruslan Salakhutdinov.
\newblock Actor-mimic: Deep multitask and transfer reinforcement learning.
\newblock \emph{arXiv preprint arXiv:1511.06342}, 2015.

\bibitem[Peng et~al.(2019)Peng, Kumar, Zhang, and Levine]{peng2019awr}
Xue~Bin Peng, Aviral Kumar, Grace Zhang, and Sergey Levine.
\newblock Advantage-weighted regression: Simple and scalable off-policy
  reinforcement learning.
\newblock \emph{arXiv preprint arXiv:1910.00177}, 2019.

\bibitem[Pong et~al.(2021)Pong, Nair, Smith, Huang, and
  Levine]{pong2021offline}
Vitchyr~H Pong, Ashvin Nair, Laura Smith, Catherine Huang, and Sergey Levine.
\newblock Offline meta-reinforcement learning with online self-supervision.
\newblock \emph{arXiv preprint arXiv:2107.03974}, 2021.

\bibitem[Radosavovic et~al.(2022)Radosavovic, Xiao, James, Abbeel, Malik, and
  Darrell]{radosavovic2022real}
Ilija Radosavovic, Tete Xiao, Stephen James, Pieter Abbeel, Jitendra Malik, and
  Trevor Darrell.
\newblock Real-world robot learning with masked visual pre-training.
\newblock \emph{arXiv preprint arXiv:2210.03109}, 2022.

\bibitem[Shafiullah et~al.(2022)Shafiullah, Cui, Altanzaya, and
  Pinto]{shafiullah2022behavior}
Nur Muhammad~Mahi Shafiullah, Zichen~Jeff Cui, Ariuntuya Altanzaya, and Lerrel
  Pinto.
\newblock Behavior transformers: Cloning $ k $ modes with one stone.
\newblock \emph{arXiv preprint arXiv:2206.11251}, 2022.

\bibitem[Siegel et~al.(2020)Siegel, Springenberg, Berkenkamp, Abdolmaleki,
  Neunert, Lampe, Hafner, and Riedmiller]{siegel2020keep}
Noah~Y Siegel, Jost~Tobias Springenberg, Felix Berkenkamp, Abbas Abdolmaleki,
  Michael Neunert, Thomas Lampe, Roland Hafner, and Martin Riedmiller.
\newblock Keep doing what worked: Behavioral modelling priors for offline
  reinforcement learning.
\newblock \emph{arXiv preprint arXiv:2002.08396}, 2020.

\bibitem[Singh et~al.(2019)Singh, Yang, Hartikainen, Finn, and
  Levine]{singh2019}
Avi Singh, Larry Yang, Kristian Hartikainen, Chelsea Finn, and Sergey Levine.
\newblock End-to-end robotic reinforcement learning without reward engineering.
\newblock \emph{Robotics: Science and Systems}, 2019.

\bibitem[Singh et~al.(2020)Singh, Yu, Yang, Zhang, Kumar, and
  Levine]{singh2020cog}
Avi Singh, Albert Yu, Jonathan Yang, Jesse Zhang, Aviral Kumar, and Sergey
  Levine.
\newblock Cog: Connecting new skills to past experience with offline
  reinforcement learning.
\newblock \emph{arXiv preprint arXiv:2010.14500}, 2020.

\bibitem[Teh et~al.(2017)Teh, Bapst, Czarnecki, Quan, Kirkpatrick, Hadsell,
  Heess, and Pascanu]{teh2017distral}
Yee~Whye Teh, Victor Bapst, Wojciech~Marian Czarnecki, John Quan, James
  Kirkpatrick, Raia Hadsell, Nicolas Heess, and Razvan Pascanu.
\newblock Distral: Robust multitask reinforcement learning.
\newblock \emph{arXiv preprint arXiv:1707.04175}, 2017.

\bibitem[Vecerik et~al.(2017)Vecerik, Hester, Scholz, Wang, Pietquin, Piot,
  Heess, Roth{\"o}rl, Lampe, and Riedmiller]{vecerik2017leveraging}
Mel Vecerik, Todd Hester, Jonathan Scholz, Fumin Wang, Olivier Pietquin, Bilal
  Piot, Nicolas Heess, Thomas Roth{\"o}rl, Thomas Lampe, and Martin Riedmiller.
\newblock Leveraging demonstrations for deep reinforcement learning on robotics
  problems with sparse rewards.
\newblock \emph{arXiv preprint arXiv:1707.08817}, 2017.

\bibitem[Wilson et~al.(2007)Wilson, Fern, Ray, and Tadepalli]{wilson2007multi}
Aaron Wilson, Alan Fern, Soumya Ray, and Prasad Tadepalli.
\newblock Multi-task reinforcement learning: a hierarchical bayesian approach.
\newblock In \emph{Proceedings of the 24th international conference on Machine
  learning}, pages 1015--1022, 2007.

\bibitem[Wu et~al.(2019)Wu, Tucker, and Nachum]{wu2019behavior}
Yifan Wu, George Tucker, and Ofir Nachum.
\newblock Behavior regularized offline reinforcement learning.
\newblock \emph{arXiv preprint arXiv:1911.11361}, 2019.

\bibitem[Wu and He(2018)]{wu2018group}
Yuxin Wu and Kaiming He.
\newblock Group normalization.
\newblock In \emph{Proceedings of the European conference on computer vision
  (ECCV)}, pages 3--19, 2018.

\bibitem[Xiao et~al.(2022)Xiao, Radosavovic, Darrell, and
  Malik]{xiao2022masked}
Tete Xiao, Ilija Radosavovic, Trevor Darrell, and Jitendra Malik.
\newblock Masked visual pre-training for motor control.
\newblock \emph{arXiv preprint arXiv:2203.06173}, 2022.

\bibitem[Xie and Finn(2021)]{xie2021lifelong}
Annie Xie and Chelsea Finn.
\newblock Lifelong robotic reinforcement learning by retaining experiences.
\newblock \emph{arXiv preprint arXiv:2109.09180}, 2021.

\bibitem[Yang and Nachum(2021)]{yang2021representation}
Mengjiao Yang and Ofir Nachum.
\newblock Representation matters: Offline pretraining for sequential decision
  making.
\newblock \emph{arXiv preprint arXiv:2102.05815}, 2021.

\bibitem[Yang et~al.(2021)Yang, Levine, and Nachum]{yang2021trail}
Mengjiao Yang, Sergey Levine, and Ofir Nachum.
\newblock Trail: Near-optimal imitation learning with suboptimal data.
\newblock \emph{arXiv preprint arXiv:2110.14770}, 2021.

\bibitem[Young et~al.(2020)Young, Gandhi, Tulsiani, Gupta, Abbeel, and
  Pinto]{young2020visual}
Sarah Young, Dhiraj Gandhi, Shubham Tulsiani, Abhinav Gupta, Pieter Abbeel, and
  Lerrel Pinto.
\newblock Visual imitation made easy, 2020.

\bibitem[Yu et~al.(2021)Yu, Kumar, Chebotar, Hausman, Levine, and
  Finn]{yu2021conservative}
Tianhe Yu, Aviral Kumar, Yevgen Chebotar, Karol Hausman, Sergey Levine, and
  Chelsea Finn.
\newblock Conservative data sharing for multi-task offline reinforcement
  learning.
\newblock \emph{NeurIPS}, 34, 2021.

\bibitem[Yu et~al.(2022)Yu, Kumar, Chebotar, Hausman, Finn, and
  Levine]{yu2022leverage}
Tianhe Yu, Aviral Kumar, Yevgen Chebotar, Karol Hausman, Chelsea Finn, and
  Sergey Levine.
\newblock How to leverage unlabeled data in offline rl.
\newblock \emph{arXiv:2202.01741}, 2022.

\end{thebibliography}
